\documentclass[conference]{IEEEtran}
\usepackage{cite}
\ifCLASSINFOpdf
  \usepackage[pdftex]{graphicx}
  \graphicspath{figures/}
\else
\fi
\usepackage{amsmath}
\ifCLASSOPTIONcompsoc
  \usepackage[caption=false,font=normalsize,labelfont=sf,textfont=sf]{subfig}
\else
  \usepackage[caption=false,font=footnotesize]{subfig}
\usepackage{colortbl}

\usepackage{verbatim}

\usepackage{enumerate}

\usepackage{booktabs}
\usepackage[table,xcdraw]{xcolor}
\usepackage{algorithm}% http://ctan.org/pkg/algorithms
\usepackage[noend]{algpseudocode}% http://ctan.org/pkg/algorithmicx
\floatname{algorithm}{\small Algorithm}

\begin{document}
\title{\textit{PEng4NN}: An Accurate Performance Estimation Engine for Efficient Automated Neural Network Architecture Search}

 %% other options
%% PEng4NN: A Performance inference Engine for efficient and accurate search of Neural Network models
%% PEng4NN: A Performance inference Engine for efficient and accurate automated search for Neural Network models
%% PEng4NN: A Performance inference Engine for efficient and accurate automated search for Neural Network architectures
%% 

% performance inference engine (pie, perfie)
%

\begin{comment}
\author{\IEEEauthorblockN{Michael Shell}
\IEEEauthorblockA{School of Electrical and\\Computer Engineering\\
Georgia Institute of Technology\\
Atlanta, Georgia 30332--0250\\
Email: http://www.michaelshell.org/contact.html}
\and
\IEEEauthorblockN{Homer Simpson}
\IEEEauthorblockA{Twentieth Century Fox\\
Springfield, USA\\
Email: homer@thesimpsons.com}
\and
\IEEEauthorblockN{James Kirk\\ and Montgomery Scott}
\IEEEauthorblockA{Starfleet Academy\\
San Francisco, California 96678--2391\\
Telephone: (800) 555--1212\\
Fax: (888) 555--1212}}
\end{comment}


\author{\IEEEauthorblockN{Ariel Keller Rorabaugh\IEEEauthorrefmark{1},
Silvina Ca\'ino-Lores\IEEEauthorrefmark{1},
Michael R. Wyatt II\IEEEauthorrefmark{1}, 
Travis Johnston\IEEEauthorrefmark{2}, and
Michela Taufer\IEEEauthorrefmark{1}}\vspace{.1cm}
\IEEEauthorblockA{\IEEEauthorrefmark{1}University of Tennessee, Knoxville, USA \IEEEauthorrefmark{2}Oak Ridge National Lab, Oak Ridge, USA \\
aror@utk.edu, scainolo@utk.edu, mike@wyatt.codes, johnstonjt@ornl.gov, taufer@acm.org}
}


\maketitle

\IEEEpeerreviewmaketitle

%% Tentative title: 
%% PEng4NN: An Accurate Performance Estimation Engine for Efficient Automated Neural Network Architecture Search

\begin{abstract}
% The abstract goes here. Max 500 words (paper is max 10 pages with everything)
% Rough draft is below---please edit it or leave comments as you like. (Initial word count: 210)

Neural network (NN) models are increasingly used in scientific simulations, artificial intelligence, and other high performance computing (HPC) fields to extract knowledge from datasets. Each dataset requires tailored NN model architecture, but designing individual structures by hand is a time-consuming and error-prone process. Neural architecture search (NAS) automates the design of NN architectures. NAS attempts to find well-performing NN models for specialized datsets, where performance is measured by key metrics that capture the NN capabilities (e.g., accuracy of classification of samples in a dataset). Existing NAS methods are still resource intensive, especially when searching for highly accurate models for larger and larger datasets. 

To address this problem, we propose a performance estimation strategy that reduces the resources for training individual NNs and increases NAS throughput without jeopardizing accuracy.
We implement our strategy via an engine called \textit{PEng4NN} that plugs into existing NAS methods; in doing so, \textit{PEng4NN} predicts the final training accuracy of NNs early in the training process, informs the NAS of NN performance, and thus enables the NAS to terminate training NNs early. We assess the effectiveness of our engine on three diverse datasets (i.e., CIFAR-100, Fashion MNIST, and SVHN). By reducing the quantity of training epochs needed, our engine achieves substantial throughput gain; on average, our engine saves 61\% to 82\% of the training epochs needed, increasing throughput by a factor of 2.5 to 5 compared to a state-of-the-art NAS method. We achieve this gain without compromising accuracy, as we demonstrate with two key outcomes. First, across all our tests, between 74\% and 97\% of the ground truth best models lie in our set of predicted best models. Second, the accuracy distributions of the ground truth best models and our predicted best models are comparable, with the mean accuracy values differing by at most .7 percentage points across all tests.

\end{abstract}

\section{Introduction}
\label{sec:intro}

Neural networks are powerful models that are increasingly leveraged in traditional HPC-oriented fields like scientific simulations and new research areas such as high-performance artificial intelligence and high-performance data analytics. However, finding a suitable neural network architecture is a time-consuming, data-dependent process involving several rounds of hyperparameter selection, training, validation, and manual inspection.

Recent works have proposed methods to automate the design of neural networks, including neural architecture search (NAS) strategies to find near-optimal models for a given dataset. NAS involves selecting neural network (NN) models from the search space, training them on the target dataset, evaluating their performance, and using information about NN performance to select new NNs from the search space. This process is computationally intensive and time-consuming for even a single network, and it is typically conducted in an iterative manner, which increases the need for computing power even further.% Nevertheless, some models are not suitable candidates to accurately classify the dataset of interest, and training the networks completely is infeasible when the goal is NN model search or NN model optimization. 

In order to improve the throughput and efficiency of NAS, NN models can be analyzed to estimate their performance and establish early termination criteria or steer searching mechanisms. Limited research exists in the area of NN performance estimation and early termination criteria, and such methods are often directly integrated into a particular NAS implementation.
In this paper, we propose \textit{PEng4NN}, a performance estimation engine that predicts fully-trained network accuracy early in the training process. Crucially, {\it PEng4NN} decouples the search and estimation strategies in state-of-the-art NAS methods: it plugs into existing NAS workflows and outputs performance estimates for the NN models. This information can be leveraged by the NAS to redirect computational resources from training, thus increasing the throughput of the NAS and allowing it to explore more candidate models. 

The contributions of this work are:
\begin{enumerate}
    \item The definition of a core accuracy predictive function that models the accuracy of an NN independently of its architecture and the dataset it aims to classify.
    \item The design of a performance estimation strategy agnostic to the NAS method and dataset of choice, capable of accurately inferring the performance of a given NN model in fewer training epochs than comparable state-of-the-art methods.
    \item The implementation of this strategy as \textit{PEng4NN}, a fully decoupled performance estimation engine that can plug into existing NAS engines for classification problems.
    \item A systematic characterization of widely accepted benchmark datasets for NN training and validation (i.e., CIFAR-100, Fashion MNIST, and SVHN).
    \item An evaluation of {\it PEng4NN} on the characterized benchmark datasets, including analysis of our engine's gain and accuracy.
\end{enumerate}

\section{Limitations of Neural Architecture Search}
\label{sec:nas}

Neural networks (NN) are powerful and flexible models that can be tailored to many complex learning tasks like speech and natural language processing, image classification, and even multidimensional scientific problems. Despite their success, NNs are still hard to design manually and they must be tuned for each specific dataset. Recent works aim to automate the design of NN for a given dataset by conducting a neural architecture search (NAS) that selects the number, type, and sequence of layers, and chooses the hyper-parameters that describe how each layer functions. In current literature there are several strategies to conduct NAS, including random searches, grid searches, hyper-parameter sweeps, reinforcement learning, evolutionary optimization, gradient-based optimization, and bayesian optimization \cite{elsken2019neural}. Most of these strategies are implemented as search engines that incorporate on-the-fly learning to improve the generated models iteratively. The aggregated load of training all the NNs in the vast search space makes NAS a computationally intensive problem that requires a high-throughput approach, especially in the early stages of NAS in which up to 88\% of NN fail to ever learn \cite{Johnston2017}. Consequently, scientists must either allocate significant time on the largest compute resources available (in the range of tens of thousands of GPU hours \cite{patton2018167}) or make assumptions about the models to simplify the search. 

 %For example, MENNDL~\cite{young2017evolving,young2019evolving} is an example of a NAS system that performs systematic search for optimum neural architecture using an evolutionary approach.

\begin{figure}[t!]
    \centering
    \includegraphics[width=.9\columnwidth]{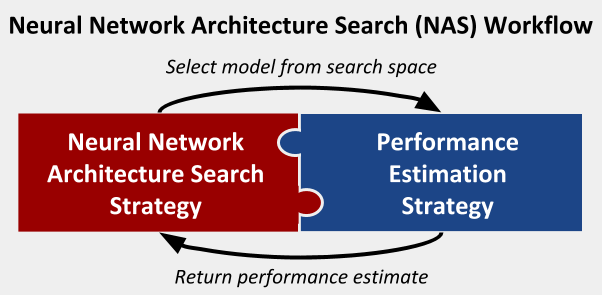}
    \caption{Overview of current NAS workflows and their tightly-coupled relationship between the search and performance estimation strategies.}
    \label{fig:nnas_workflow}
\end{figure}

To alleviate this issue, NAS can be supported by performance estimation strategies that provide information (e.g., accuracy or loss) for a given model trained on a specific dataset. These estimates enable the NAS to truncate the training process, compare partially-trained NNs, and steer the search towards models that yield better performance. As a result, performance estimation strategies can improve the efficiency and throughput of the NAS. This results in an overall NAS workflow in which the NAS strategy selects an NN model from the search space for a specific dataset and feeds this model to the performance estimation strategy, which returns a performance estimate that can be leveraged by the NAS. Figure \ref{fig:nnas_workflow} shows the tightly-coupled vision of this NAS workflow as found in the literature \cite{elsken2019neural}. This yields that a performance estimation strategy is tailored for a specific NAS method, which imposes restrictions to the generalization of the performance estimation strategy for different NAS engines. A coupled design also prevents conducting optimizations in the resource allocation for each component of the workflow, their orchestration, and the mechanisms to exchange data efficiently amongst them. All of these limitations can reduce the efficiency and throughput of the NAS workflow as a whole. This work addresses these limitations by designing a decoupled performance estimation strategy that we implement as the \textit{PEng4NN} performance estimation engine. By decoupling the NAS from \textit{PEng4NN}, we can plug our engine into diverse NAS applied to different datasets. 

%Our work introduces a performance estimation strategy that complements the NAS strategies for high-throughput network architecture search and efficiency. Our strategy is introduced in Fig.~\ref{fig:peng4nn} as the \textit{performance estimation engine}, which is built around a \textit{core accuracy predictive function}. It includes NN training and validation modules, and a prediction analyzer. The performance estimation engine helps the NAS focus the search into regions that contain the best networks. Because we can reduce the training epochs without hurting the accuracy of the best models, the NAS can conclude more quickly or use modest computational resources and still find (near) optimal models.

 %In addition, our engine is designed for state-of-the-art high-end systems, and it leverages accelerators (e.g., GPGPUs) for efficient NN training, and spare CPU resources for analysis in a manner that exploits locality and minimises data transfers. \todo{is this true right now?} 

\section{Performance Estimation Engine (\textit{PEng4NN)}}
\label{sec:cfp}

\begin{figure*}[t!]
    \centering
    \includegraphics[width=.8\textwidth]{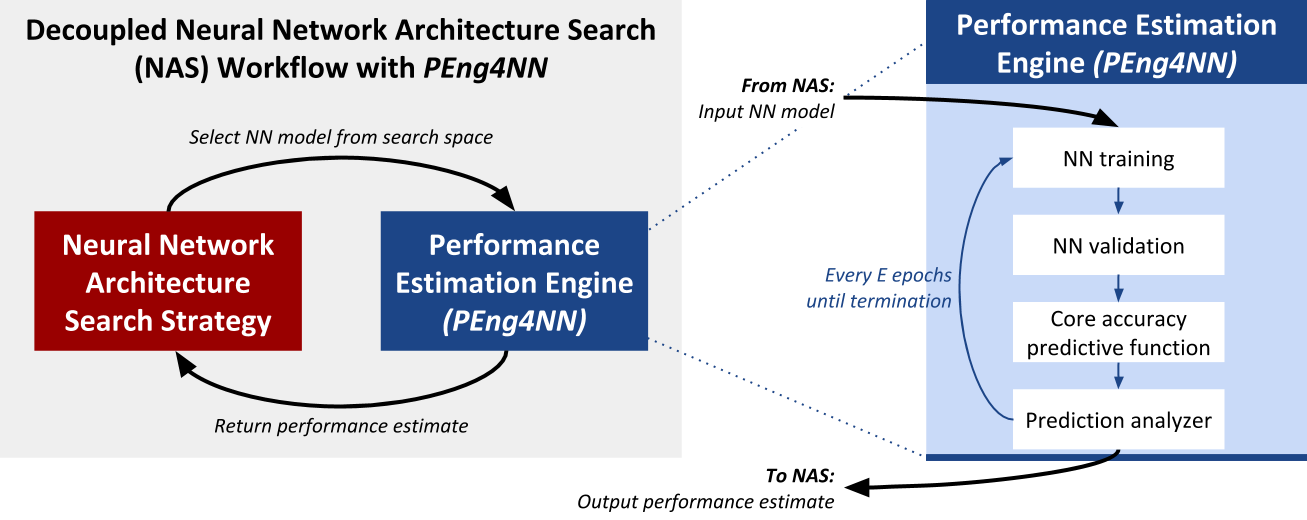}
    \caption{Overview of \textit{PEng4NN}, our solution of a decoupled performance estimation engine that, given an input NN model provided by the NAS, outputs a performance estimate that can be leveraged by the NAS.}
    \label{fig:peng4nn}
\end{figure*}

Our performance estimation strategy relies on a
fully decoupled NAS and NN performance estimation, as shown in Figure~\ref{fig:peng4nn}. In doing so, we pursue a strategy that is agnostic to  the  NAS  methods  and  datasets. We implement this strategy in \textit{PEng4NN}, our fully decoupled performance estimation engine. \textit{PEng4NN} augments NAS strategies to improve their throughput and efficiency. Each time the NAS selects an NN model, it first passes that model to \textit{PEng4NN}. The performance estimation engine iteratively executes a four-step process depicted in Figure~\ref{fig:peng4nn}. \textit{PEng4NN} begins each iteration by training and validating the NN. It then constructs a core accuracy predictive function based on the NN's validation performance. \textit{PEng4NN} passes the prediction derived from this function to the prediction analyzer that determines whether a stable performance estimate was found. With this information, \textit{PEng4NN} decides whether to output a final estimate or continue the iterative process.
The following subsections describe in detail the components of \textit{PEng4NN}.

\subsection{NN training and validation}\label{sec:train-val}

The dataset of interest for the classification problem is divided into a training set and a validation set. At each iteration, \textit{PEng4NN} trains a given NN model $M$ on the training data for a constant number of epochs, \(E\), where an epoch means one cycle through the training set. The value of parameter \(E\) is user-defined. In all of our tests, we let \(E = 0.5\), meaning that \textit{PEng4NN} trains the NN for 0.5 epochs at each iteration (i.e., the NN is trained on half the training data). \textit{PEng4NN} then validates the NN and calculates its classification accuracy ($a_V$) and loss ($l_V$) for the validation set at the current epoch ($e$). Note that we select a batchsize and truncate our datasets to be divisible by the batchsize. As a result we ensure we can train for precisely 0.5 epochs. 

%Note that an epoch means one cycle through the training set; depending on divisibility of the training set size, a training set cycle may involve training on a subset of the training set, with a constant small number of images from the training set randomly left out each epoch cycle. 

\subsection{Core accuracy predictive function}

%The epoch and newly calculated performance measurements (i.e., validation accuracy and loss) from the training and validation of the current iteration are added to an ordered list of epoch and accuracy datapoints for the NN ($V$); a core accuracy predictive function is fitted to the NN's epoch and accuracy data. This function gives a maximum accuracy prediction ($a_P$) for the NN at a given epoch $e$.

We define the predictive function that we fit to the NN accuracy data across epochs, and we analytically determine bounds and initial values for the predictive function’s parameters. Then, we describe how we leverage Sci-Py’s \textit{optimize.curve\_fit} method (Curve Fit) to fit our chosen function to the historic accuracy data of a given NN model. Finally, we show how we use the fitted predictive function to determine a maximum accuracy prediction for the NN at each iteration of the engine.

\subsubsection{Defining the function}

\begin{figure}[!tbp]
\begin{center}
    \subfloat[\(f(x)=32-b^{(5-x)}\)]{\includegraphics[trim=12 13 12 13, clip,width=.75\linewidth]{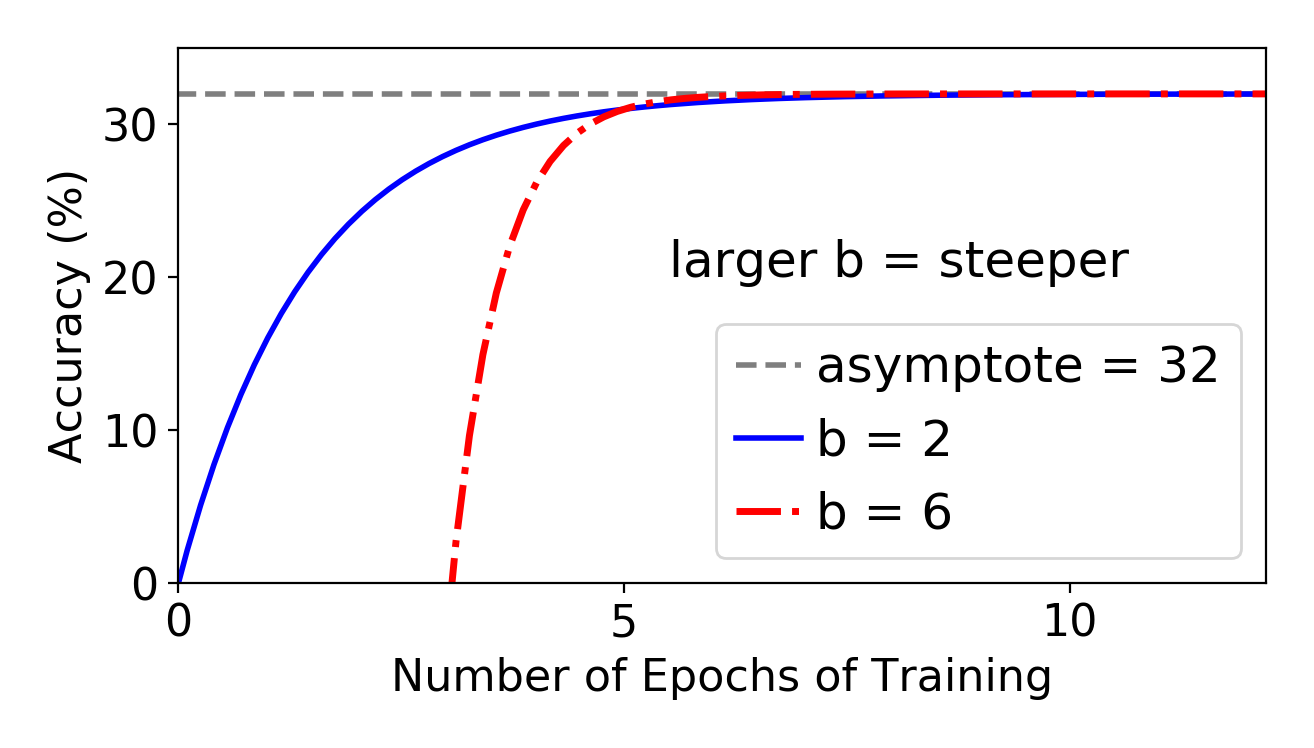}\label{fig:parameterB}} \\
    \subfloat[\(f(x)=32-2^{(c-x)}\)]{\includegraphics[trim=12 13 12 13, clip,width=.75\linewidth]{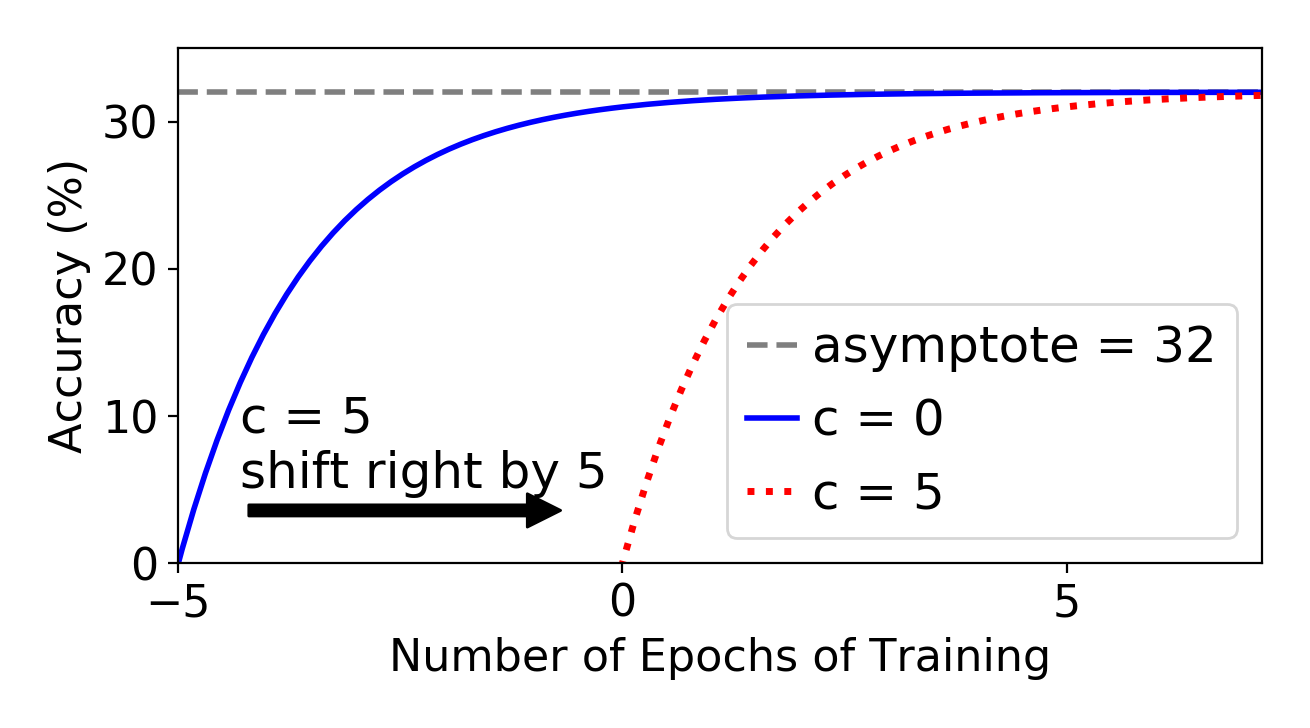}\label{fig:parameterC}}
    \caption{Sample function with different values for parameters \(b\) and \(c\)}
    \label{fig:fnParameters}
\end{center}
\end{figure}

We hypothesize that NN accuracy curves can be modeled with functions of the form \(f(x)=a - b^{(c-x)}\). 
Empirical observation suggests that the accuracy curves of NNs tend to be concave down and increasing, with a horizontal asymptote that the accuracy values approach; the family of functions \(f(x)=a - b^{(c-x)}\) shares these properties. The function parameter \(a\) describes the horizontal asymptote; the parameter \(b\) controls steepness of the curve/corner; the parameter \(c\) denotes a horizontal shift. Figure \ref{fig:parameterB} depicts a sample function with different values for \(b\): \(f(x)=32-b^{(5-x)}\), with a solid blue line for \(b=2\) and a dashed red line for \(b=6\). The dashed red line with the larger \(b\) value is the steeper curve, illustrating that larger values for the parameter \(b\) correspond to steeper curves.
Figure \ref{fig:parameterC} depicts a sample function with different values for \(c\): \(f(x)=32-2^{(c-x)}\), with a solid blue line for \(c=0\) and a dotted red line for \(c=5\). The dotted red line with \(c=5\) is shifted right by 5 points as compared to \(c=0\), illustrating that the parameter \(c\) denotes a horizontal shift.

\subsubsection{Bounding and initializing the function}

We define bounds for the three function parameters \(a\), \(b\)\, and \(c\). 

% \paragraph{Bounding \(a\)}
{\it Bounding \(a\):}
The maximum accuracy prediction (defined by parameter \(a\)) should be strictly greater than 0\% and at most 100\%. In many cases, the accuracy of an NN increases very quickly at first, and the rate of improvement decreases over epochs. 

During early iterations of \textit{PEng4NN} Curve Fit has limited datapoints for fitting the function. We observe empirically for some NNs that during these first steps Curve Fit expects unreasonable accuracy growth because it sees only the steepest portion of the accuracy curve with the largest rate of increase. Then, if \(a\) is unbounded, Curve Fit selects a larger value for \(a\) than is reasonable (i.e., larger than 100\%). We want to distinguish this scenario from NNs that are simply very accurate. If we bound \(a\) by 100, then Curve Fit may predict 100\% accuracy in both situations described above.
Instead, we allow \(a\) to be slightly larger than 100. This enables us to disallow convergence if the value of \(a\) is greater than 100 (i.e., in the case that Curve Fit has seen only the steepest part of the curve and predicts unreasonable accuracy growth). %In {\it PEng4NN} the bounds we assign for \(a\) are: \(0.5 \leq a \leq 102.5\)
In {\it PEng4NN}, we bound \(a\) by \(0.5 \leq a \leq 102.5\) and initialize it with \(a = 10\).

% \paragraph{Bounding \(b\)}
{\it Bounding \(b\):}
The function is undefined at any non integer powers if \(b<0\); therefore, \(b\) must be strictly positive.
Furthermore, if \(b<1\), then the function no longer has the desired shape; instead of increasing up to a horizontal asymptote as \(x\) increases, the function approaches negative infinity as \(x\) increases. 
Thus, \(b \geq 1\). In {\it PEng4NN}, we bound \(b\) by \(1 \leq b \leq \infty\) and initialize it with \(b = 1.001\).

% \paragraph{Bounding \(c\)}
{\it Bounding \(c\):}
The parameter \(c\) represents a horizontal shift of the function.
We begin by proving that any horizontal shift to the left (that is, a shift by \(c\) where \(c\) is negative) implies that \(f(1)\geq a-1\).

Consider the function \(f(x)=a-b^{(c-x)}\).
If \(c < 0\), it follows that \(f(1)=a-b^{i}\), where \(i < -1\), or \(f(1)=a-\frac{1}{b^{j}}\), where \(j>1\). Then, because \(b \geq 1\), it follows that \(f(1)\geq a-1\).

Before passing the validation accuracy values per epoch to Curve Fit, we rescale the \(x\) values so that the initial value for \(x\) is 1, meaning that the corresponding epoch values will be rescaled to start at 1.
Then, \(f(1)\geq a-1\) means that the initial accuracy is within 1\% of the maximum accuracy; in other words, the NN accuracy improves by only 1\% throughout training.
Such behavior is modeled by a flat line with \(b=1\), in which case the value of \(c\) has no effect on the function. Because of this observation, we conclude that horizontal shifts to the left are not needed, and hence \(c \geq 0\). In {\it PEng4NN}, we bound \(c\) by \(0 \leq c \leq \infty\) and initialize it with \(c=100\).

%Given these bounds, we select initial values for \(a\), \(b\), \(c\) that lie within the bounds derived above: \(a=10\), \(b=1.001\), \(c=100\).

\subsubsection{Fitting the function to predict accuracy}

Algorithm \ref{alg:core} shows the internal procedure in the core accuracy predictive function. Given an NN model $M$, the procedure takes as input the current training epoch ($e$), the history of validation performance and predicted accuracy as a list ordered by epoch ($\mathcal{H}$), the validation accuracy ($a_V$) and loss ($l_V$) at epoch $e$, and the predictive function \(f(x)=a-b^{(c-x)}\) that is bounded and initialized with the values specified in the previous section. 

The first step in our core predictive procedure is the creation of a new tuple of performance metrics for $M$ at the specified epoch ($h_e$). This tuple is inserted in the performance history $\mathcal{H}$ for $M$ (Line \ref{fun:insert}) so that it is taken into consideration as a new datapoint for the Curve Fit method. The rest of the steps required to fit the function $f$ only take place if there are enough datapoints to conduct the Curve Fit method. In this case $f$ is an exponential function, thus we need at least $C_{min}=3$ datapoints in $\mathcal{H}$ to proceed (Line \ref{if:cardinality}).

Recall that the constant number of training epochs per iteration of \textit{PEng4NN} (\(E\)) is user-defined (see Sec.~\ref{sec:train-val}). We rescale $\mathcal{H}$ as a new list $\mathcal{H}'$ in which the first epoch value is 1 (Line \ref{fun:rescale_epochs}), since this enables us to assign general bounds for the parameters \(a\), \(b\), and \(c\) that do not depend on the value of \(E\).  For example, if the NN is validated every half epoch, the first epoch value is 0.5, so $\mathcal{H}$ is rescaled by multiplying by 2 every value of $e$ in the list. Then we feed $\mathcal{H}'$ to SciPy's \textit{optimize.curve\_fit} method to find the function \(f\) that best fits the historic accuracy data for the NN model (Line \ref{fun:curve_fit}). We define the maximum accuracy prediction $a_P$ given by the predictive function \(f\) as the value of the parameter \(a\) that results from Curve Fit. We update this value $a_P$ in the performance tuple for the current epoch ($h_e$) so that it is reflected in the performance history (Line \ref{fun:update}).
% Curve Fit uses least squares to determine the values for parameters \(a\), \(b\), and \(c\) that best fit the accuracy data in $\mathcal{H}'$.

As a result of the core accuracy predictive function, we output the updated history of validation performance and accuracy predictions ($\mathcal{H}$) for further analysis.
%We pass to Curve Fit the following set of arguments:
%\begin{itemize}
%    \item the rescaled epoch data and accuracy data,
%    \item the predictive function defined above, and
%    \item the parameter bounds and initial values described %below.
%\end{itemize}

%The maximum accuracy prediction derived from the core accuracy predictive function is appended to an ordered list of all the max accuracy predictions for the NN, and this list is given to the prediction analyzer. 

\subsection{Prediction analyzer}
%Every time a new prediction is made, it is added to the list of all predictions for the given NN, and the updated list is passed to the stabilizer.

Each time the core accuracy predictive function receives a new validation performance measurement for an NN model $M$ at epoch $e$, it generates a maximum accuracy prediction $a_P$ that is appended to the history of measured and predicted values $\mathcal{H}$. The fitted function and the associated $a_P$ vary from one iteration to the next because new performance measurements are appended to $\mathcal{H}$. The goal of the prediction analyzer is to determine whether the performance estimates for $M$ have converged to a stable value, which we denote by $A_P$. %If so, the analyzer outputs the performance estimate $A_P$ and \textit{PEng4NN}'s iterative process terminates.% If not, the next iteration begins and the NN model resumes training. << im getting rid of this because it's more complicated thann that

\begin{table}[t!]
% \caption{Symbol summary for the core accuracy predictive function component of \textit{PEng4NN}.}
\caption{Symbols for accuracy predictive function algorithm.}
\begin{center}
\label{tab:symb_core}
\begin{tabular}{@{}rp{.85\columnwidth}@{}}
\toprule[2pt]
\textbf{Symbol} & \textbf{Definition} \\ 
\midrule[1pt]
$M$ & NN model identifier \\ \midrule[.5pt]
$e$ & Current epoch in which the prediction will be calculated \\ \midrule[.5pt]
$a_{V}$ & Validation accuracy from \textit{NN validation} component \\ \midrule[.5pt]
$l_{V}$ & Validation loss from \textit{NN validation} component \\ \midrule[.5pt]
$a_{P}$ & Predicted accuracy  \\ \midrule[.5pt]
$\mathcal{H}$ & Ordered list of $<M, e, a_{V}, l_{V}, a_{P}>$ tuples, each representing the validation accuracy $a_V$ and loss $l_V$ of an NN $M$ at a specific epoch $e$, plus the accuracy prediction $a_P$ \\ \midrule[.5pt]
$f$ & Initial function for the Curve Fit method \\ \midrule[.5pt]
$C_{min}$ & Minimum cardinality of the dataset $f$ will be fitted to \\ \midrule[.5pt]
$a$ & Horizontal asymptote for the fitted function at epoch $e$\\ \midrule[.5pt]
$b$ & Steepness of the fitted function at epoch $e$ \\ \midrule[.5pt]
$c$ & Horizontal shift of the fitted function at epoch $e$\\ 
\bottomrule[2pt]
\end{tabular}
\end{center}
%\textsuperscript{\textdagger}\footnotesize{SVHN also contains 531,131 samples that can be used as additional training data, which we did not use in our evaluation.}
\end{table}

\begin{algorithm}[t]
  \caption{ \small Procedure for accuracy predictive function.}\label{alg:core}
%   \caption{Procedure for the core accuracy predictive function. See Tab.~\ref{tab:symb_core} for symbol definitions.}\label{alg:core}
  \begin{algorithmic}[1]
    \small
    \Require{\small Current training epoch ($e$); history of validation performance and predicted accuracy ($\mathcal{H}$); validation performance at epoch $e$ (i.e., accuracy, $a_V$, and loss, $l_V$); function to fit ($f$) and minimum required cardinality ($C_{min}$)}
    \Ensure{\small History of validation performance and predicted accuracy ($\mathcal{H}$), updated with the latest prediction $a_P$}
    \Procedure{Predictor}{$e$, $\mathcal{H}$, $a_V$, $l_V$, $f$}
    \State $h_e \gets <M, e, a_V, l_V, 0>$
    \State $\mathcal{H} \gets insert(h_{e})$\label{fun:insert}
    \If{$|\mathcal{H}| \geq C_{min}$} \label{if:cardinality}
      \State $\mathcal{H}'\gets rescale\_epochs(\mathcal{H})$\label{fun:rescale_epochs} %\Comment{Epochs must start in 1 to respect analytical bounds}
      \State $a,b,c\gets optimize.curve\_fit(\mathcal{H}', f)$ \label{fun:curve_fit}
      \State $a_{P}\gets a$ %\Comment{The predicted accuracy is the parameter $a$ in the fitted function} 
      \State $\mathcal{H} \gets update\_prediction(\mathcal{H}, h_e, a_{P})$\label{fun:update}
    \EndIf
    \State \textbf{return} $\mathcal{H}$
    \EndProcedure
    \normalsize
  \end{algorithmic}
\end{algorithm}
We describe the prediction analyzer component of \textit{PEng4NN} by first defining the parameters that control the prediction analysis. We then describe the conditions to detect convergence and show how we compute the output estimate. 
%Thus, the  During early iterations, there can be large variability between these predictions. However, once there is enough data for the fitted predictive function to identify the shape of the accuracy curve, the new datapoints already match the expected shape. Hence, the predictive function and associated max accuracy prediction vary less across iterations. For a given dataset, this point at which predictions converge changes depending on the neural network. Our prediction analyzer programmatically detects when the last few maximum accuracy predictions have small variability. At that point, the estimation engine stops the iterative prediction process and reports a final performance estimate for the NN.

\subsubsection{Parameterizing the analysis}\label{sec:parameterize-analysis}

We determine six parameters to control the behavior of the prediction analyzer and its convergence criteria:
\begin{itemize}
\item \(N\), the number of most recent accuracy predictions to consider. We use \(N=3\). 
\item \(E\), the number of epochs per iteration. We use \(E=0.5\). 
\item \(e_{max}\), the maximum number of epochs we allow an NN to train. We use \(e_{max}=20\). 
\item \(t\), the threshold describing how much variability is small enough to establish convergence and report a final performance estimate. We use \(t=0.5\). 
\item \(LOSS\), a boolean describing whether or not to enable a validation loss check in the analyzer. It is useful for datasets on which an NN's accuracy may be noisy or slow to increase, more commonly encountered in unbalanced datasets. We use \(LOSS=\) \textit{true} for unbalanced datasets and \(LOSS=\) \textit{false} otherwise. When the loss check is enabled, it is only activated if the latest accuracy prediction indicates that the given NN is not performing better than guessing or has not learned more than a single class (for example, accuracy of 10\% in the case of a balanced 10 class dataset). We call such an accuracy prediction a \textit{never-learn prediction}.
\item \(L\), the number epochs to consider in the loss check. We use \(L=5\).
\end{itemize}
These parameters, summarized in Tab.~\ref{tab:symb_analyzer}, constitute part of the input arguments for the prediction analyzer procedure and can be adjusted in \textit{PEng4NN} to tune the overall iterative process. %to reflect specific characteristics of the dataset we want to optimize the NNs for. We provide a comprehensive discussion in Sec.~\ref{sec:eval}. \todo{i need feedback with this. can we support this? it would be nice!}

\subsubsection{Analyzing the convergence of the predictions}

Algorithm \ref{alg:analyzer} shows the internal procedure in the prediction analyzer. Given an NN model $M$, the procedure takes as input the current training epoch ($e$), the history of validation performance and predicted accuracy ($\mathcal{H}$), and the six configuration parameters described above.
% that checks for the convergence of the accuracy predictions generated by the core accuracy predictive function
We first check if we have reached the minimum number of epochs to assess convergence (Line \ref{line:min_epochs}). If not, this means the number of tuples in $\mathcal{H}$ is not sufficient to analyze whether the predictions are stable, so we must continue training. On the other hand, if we reached the maximum number of training epochs ($e_{max}$), we return the maximum observed validation accuracy $a_V$ (Line \ref{line:max_epochs}) with convergence status \textit{false}.
%During each iteration of the estimation engine, a new maximum accuracy prediction is made. This prediction is added to an ordered list of all predictions for the given NN, and the updated list is passed to the prediction analyzer. The prediction analyzer determines whether to start a new iteration or whether to stop the iterative process and report a final performance estimate. 

We assume that, if convergence is achieved, the final performance estimate ($A_P$) corresponds to the accuracy predicted by the predictive function in the current epoch ($a_P$) (Line \ref{line:estimate}). We define three conditions for reaching convergence:
\begin{itemize}
\item \label{analyticsitm:less100} \textit{Condition 1:} The most recent maximum accuracy prediction ($A_P$) must be less than or equal to 100\%. Otherwise, the prediction is not valid to establish convergence (Line \ref{if:less100}). Recall that parameter \(a\) of the fitted function determines the max accuracy prediction. Recall also that we allow Curve Fit to choose \(a\) larger than 100 so that we can detect if Curve Fit is predicting unreasonable accuracy growth. This condition ensures that we have a realistic value for $A_P$.
\item \label{analyticsitm:mean} \textit{Condition 2:}  The \(N\) most recent maximum accuracy predictions are all within the threshold \(t\) of their mean (Line \ref{if:mean}). In other words, for each of the \(N\) most recent predictions \(p_i\), it checks if \(mean - t \leq p_i \leq mean + t\).
\item \label{analyticsitm:losscheck} \textit{Condition 3:} If the loss check is activated (Line \ref{if:losscheck}) and the most recent maximum accuracy prediction $a_P$ is a never-learn prediction (Line \ref{if:neverlearn}), then the minimum validation loss $l_{V_{min}} = min\{l_V\in\mathcal{H}\}$ has not decreased for at least $L$ epochs (Line \ref{if:minimum}). This condition ensures that the NN is not merely learning slowly or irregularly, as can be the case especially with unbalanced datasets. On the contrary, if the accuracy is not increasing, and the validation loss is not decreasing, this indicates that the NN is likely not continuing to learn.
\end{itemize}

%Finally, if the analytics loss check is enabled, and if the latest accuracy prediction indicates that the NN is not performing better than guessing, or has not learned more than a single class, (for example, accuracy of 10\% in the case of a 10 class dataset) then we also check:
%\begin{itemize}
%\item Has the validation loss not decreased for at least 5 epochs?
%\end{itemize}

\begin{table}[t!]
\caption{Symbols for prediction analyzer algorithm.}
% \caption{Symbol summary for the prediction analyzer component of \textit{PEng4NN}.}
\begin{center}
\label{tab:symb_analyzer}
\begin{tabular}{@{}rp{.85\columnwidth}@{}}
\toprule[2pt]
\textbf{Symbol} & \textbf{Definition} \\ 
\midrule[1pt]
$e$ & Current epoch in which the predictions are analyzed  \\ \midrule[.5pt]
$\mathcal{H}$ & Ordered list of $<M, e, a_{V}, l_{V}, a_{P}>$ tuples, each representing the validation accuracy $a_V$ and loss $l_V$ of an NN $M$ at a specific epoch $e$, plus the accuracy prediction $a_P$ \\ \midrule[.5pt]
$N$ & Number of most recent accuracy predictions to consider in the analysis. We use $N=3$ \\ \midrule[.5pt]
$E$ & Number of epochs per iteration. We use $E=0.5$ \\ \midrule[.5pt]
$e_{max}$ & Maximum number of epochs to train. We use $e_{max}=20$ \\ \midrule[.5pt]
$t$ & Maximum performance variability allowed for convergence. We use $t=0.5$ \\ \midrule[.5pt]
$LOSS$ & Indicates whether the loss check is enabled \\ \midrule[.5pt]
$L$ & Number of epochs to consider in the loss check. We use $L=5$ \\ \midrule[.5pt]
$l_{V_{min}}$ & Minimum validation loss observed \\ \midrule[.5pt]
$A_{P}$ & Most recent accuracy estimate for the NN model $M$ on convergence \\
\bottomrule[2pt]
\end{tabular}
\end{center}
\end{table}

\begin{algorithm}[t]
 \caption{\small Procedure for prediction analyzer.}\label{alg:analyzer}
%   \caption{Procedure for the prediction analyzer. See Tab.~\ref{tab:symb_analyzer} for symbol definitions.}\label{alg:analyzer}
  \begin{algorithmic}[1]
    \small
    \Require{\small Current training epoch ($e$); history of validation performance and predicted accuracy ($\mathcal{H}$); configuration parameters}
    \Ensure{\small Convergence status (\textit{converged})}
    \Ensure{\small Performance estimate (predicted accuracy ($A_P$) if convergence is achieved; maximum observed validation accuracy ($a_V$) if $e_{max}$ is reached)}
    \Procedure{Analyzer}{$e$, $\mathcal{H}$, $N$, $E$, $e_{max}$, $t$, $LOSS$, $L$}
      \State \textit{converged} $\gets$ \textit{false}
      \If {$e \leq N\cdot E$} \label{line:min_epochs} %\Comment{we don't have enough datapoints/tuples}
        \textbf{exit}
      \EndIf
      \If{$e=e_{max}$} \label{line:max_epochs} %\Comment{we couldnt converge in the maximum number of epochs}
        \Return $max\{a_{V}\in \mathcal{H}\}$ 
      \EndIf
      
      \State $A_{P} \gets a_{P}$ from $\mathcal{H}$ in epoch $e$ \label{line:estimate}

      \If {$A_{P} > 100$} \label{if:less100} %prediction not valid
        \textbf{exit}
      \EndIf
      
      \If {last $N$ $a_V\in\mathcal{H}$ are within $t$} \label{if:mean} 
        \If {$LOSS$ is \textit{true}}\label{if:losscheck}
          \If{$A_{P}$ is never-learn} \label{if:neverlearn} 
             \If {last $L\cdot E$ $l_V\in\mathcal{H} \geq l_{V_{min}}$} \label{if:minimum}
               \State \textit{converged} $\gets$ \textit{true}
             \EndIf
          \EndIf
        \Else~\textit{converged} $\gets$ \textit{true}
        \EndIf
      \EndIf
      \If {\textit{converged} is \textit{true}}
        \Return $A_P$ \label{line:finalestimate}
      \Else~\textbf{exit} \label{line:end}
      \EndIf
    \EndProcedure
    \normalsize
  \end{algorithmic}
\end{algorithm}

There are two possible outcomes based on these conditions. If the conditions are all true, it means the prediction has converged. In this case, the performance estimation engine outputs $A_P$ with convergence status \textit{true} (Line \ref{line:finalestimate}). On the other hand, if any one of the conditions is not met, and the NN has not yet trained for $e_{max}$ epochs, the procedure ends with convergence status \textit{false} (Line \ref{line:end}).

Based on the output of the prediction analyzer, we can establish the termination criteria for \textit{PEng4NN}. If a performance estimate is returned, \textit{PEng4NN} terminates and sends this value to the NAS. This can occur in two scenarios: either we reached convergence and the output is $A_P$, or we reached $e_{max}$ and the output is $max\{a_V\in\mathcal{H}\}$. We can distinguish these cases by inspecting the value of the convergence status. On the other hand, if no value is returned, then \textit{PEng4NN} will begin the next iteration and resume training the NN model.

\begin{figure}[!t]
\begin{center}
    \includegraphics[trim=9 10 11 10, clip, width=.8\linewidth]{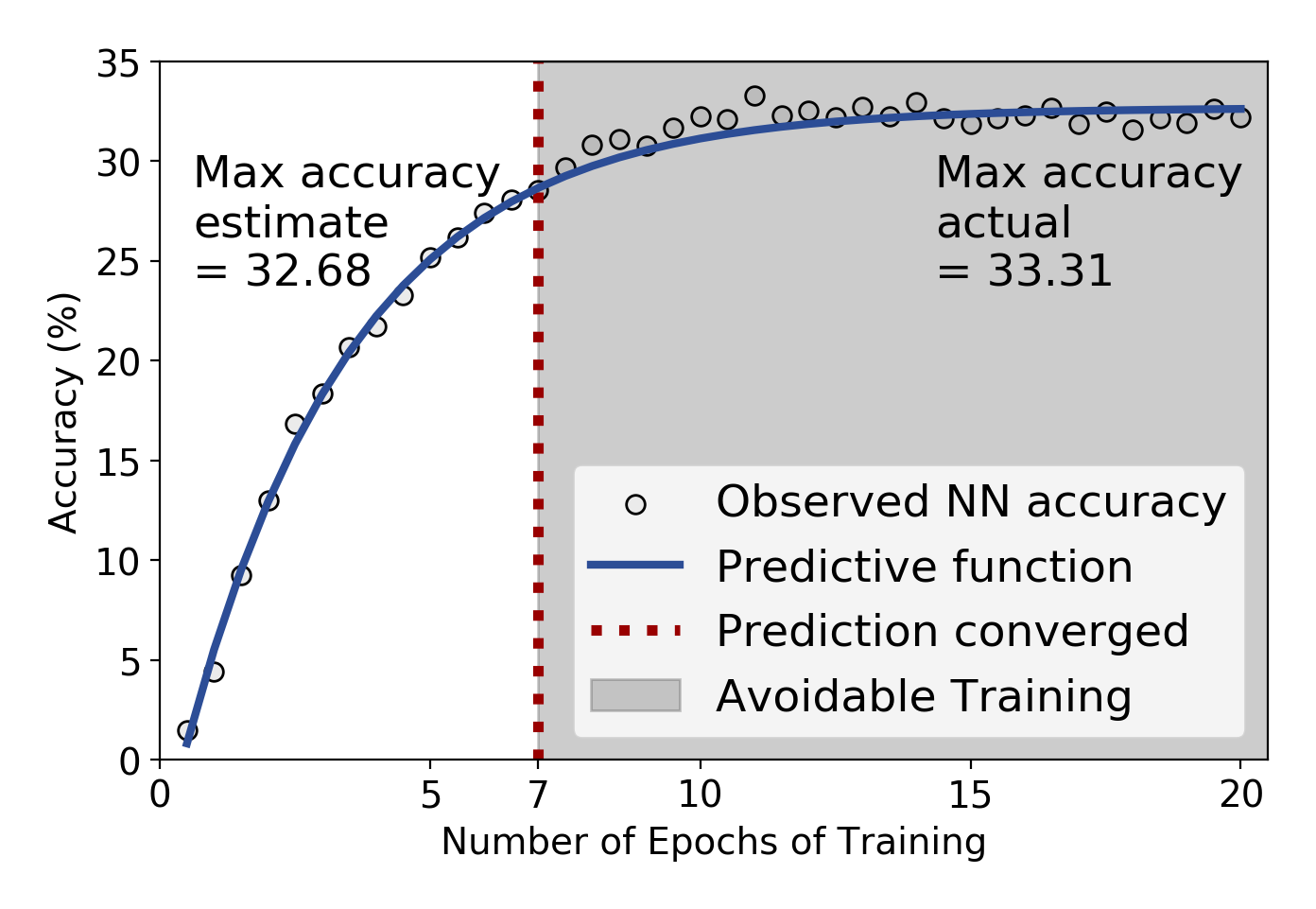}
    \caption{Example of the estimate and predictive function given by \textit{PEng4NN} for a neural network trained on CIFAR-100}
    \label{fig:cfp_example}
\end{center}
\end{figure}

As an example, Figure \ref{fig:cfp_example} depicts the end result of this process for one of the NNs we trained on CIFAR-100. The x-axis indicates the number of training epochs, and the y-axis indicates accuracy of the NN. The dotted red line indicates the point at which the prediction converges; this is when our engine outputs the performance estimate and terminates the iterative training process. In this example, our engine terminates the iterative process and outputs the performance estimate after 7 epochs. The predictive function from the final iteration is graphed, along with validation accuracy datapoints for the NN across $e_{max}=20$ epochs for comparison. Note that only the accuracy values from the first 7 epochs were used to fit the predictive function and estimate performance.

\section{Evaluation}
\label{sec:eval}

%We use techniques similar to those we employed in~\cite{JohnstonICCS} to predict the trained accuracy of randomly generated NNs being trained on datasets with diverse characteristics. 
%We begin by describing our experimental setup, including a characterization of the datasets used as benchmarks and a description of the hardware and software settings. 
We describe our experimental setup and evaluate our engine to answer two key questions: (i) \textit{``What savings are gained by our approach?"} and (ii) \textit{``How accurate are the predictions made by our approach?"}

%In answer to the first question we discuss the percentage of training epochs saved using Curve Fit Predict as compared to MENNDL's early termination criterion, and we demonstrate how this savings yields a theoretical increase in throughput. To answer the second question, we study two metrics: \textit{overlap}, which represents the proportion of the ground truth best models that are also in the set of predicted best models, and \textit{accuracy distribution}, which depicts the similarity between the neural network accuracies of the ground truth best models and predicted best models.

\subsection{Experimental Setup}

%Our evaluation uses three datasets that are widely accepted as benchmarks in the ML community: CIFAR-100, Fashion MNIST, and SVHN.

\subsubsection{Dataset Characterization} \label{sec:datasets}
To the best of our knowledge, there is no systematic characterization of datasets used in ML evaluations. This makes the assessment of the generality and applicability of a method in the area of NN research a case-driven discussion \cite{goodfellow2013maxout,lee2015deeply,shi2016improving,song2018mat,han2018batch,li2018deep} that is also found in works tackling the problem of evolutionary NN design \cite{assunccao2019denser,tan2020accurate}. We characterize three of the most commonly used datasets (i.e., CIFAR-100, F-MNIST, and SVHN) that act as benchmarks in the evaluation of novel methods and techniques for NN.
Four key attributes demonstrate the diverse nature of these datasets---number of samples, number of classes, number of image color channels, and sample size \cite{bilalli2017predictive}, plus an additional aspect of the internal balancing of the sample distribution per class, which is known to affect the performance of NNs in terms of accuracy \cite{johnson2019survey}. Table~\ref{tab:datasets} summarises these attributes and shows a sample from each dataset. We evaluate \textit{PEng4NN} with these datasets that we describe below.

%, although there are efforts towards them in the scientific community \cite{van2013openml}.

%This is particularly relevant for the \textit{meta-learning} problem, which seeks to predict the perfomance (e.g., predictive accuracies, execution time, etc.) of an ML algorithm on a given dataset, based on a set of \textit{meta-features} that capture relevant characteristics of the dataset \cite{pavel2002decision}. The definition and extraction of said meta-features is an open problem out of the scope of this work.

%efforts towards meta-learning here, stressing the need to extract meta-features from datasets as a complex task yet not solved in its reasearch area (and out of scope of this paper becuase we attempt to estimate accuracy training directly in the entire dataset, not just a set of meta-features)

{\it CIFAR-100:} This dataset was introduced in 2009 \cite{krizhevsky2009learning} as a subset of the \textit{80 Million Tiny Images} dataset \cite{torralba200880}, aimed towards improving tasks of unsupervised training of deep generative models. It is still one of the most popular benchmarks in the field of computer vision due to the manageable size of the dataset, the resolution of its images, and its challenges for NN models~\cite{darlow2018cinic,scheidegger2020efficient, barz2020we}.
%, which reduce the training time of CNNs. 
% It is often used to evaluate novel methods and model architectures in the field of deep learning \cite{barz2020we}, since it is arguably considered one of the more difficult datasets to challenge deep learning models with \cite{darlow2018cinic,scheidegger2020efficient}.

{\it Fashion MNIST (F-MNIST):} This dataset \cite{xiao2017fashion} serves as a replacement for the original MNIST dataset comprising ten classes of handwritten digits \cite{lecun1998gradient}. It shares the same image and dataset size, data format, and structure of training and testing splits with MNIST, making it a popular benchmark for NN models targeting computer vision problems \cite{meshkini2019analysis,kayed2020classification,bhatnagar2017classification}.

{\it SVHN:} Like MNIST, SVHN \cite{netzer2011reading} contains digits, but in this case they are obtained from real-world house numbers, thus contain color information, various natural backgrounds, overlapping digits, and other distracting features. These characteristics make SVHN a more difficult dataset compared to MNIST, which makes SVHN a very popular benchmark for NNs \cite{sermanet2012convolutional}. 

\begin{comment}
\begin{figure}[tbp]
\begin{center}
    \subfloat[CIFAR-100 \cite{krizhevsky2009learning}]{\includegraphics[ width=\linewidth]{figures/CIFAR100_example_data.png}\label{fig:images-CIFAR}}
    
    \subfloat[F-MNIST \cite{xiao2017fashion}]{\includegraphics[ width=\linewidth]{figures/FashionMNIST_example_data.png}\label{fig:images-FashionMNIST}}
    
    \subfloat[SVHN \cite{netzer2011reading}]{\includegraphics[ width=\linewidth]{figures/SVHN_example_data.png}\label{fig:images-SVHN}}
    \caption{Sample images from each of our benchmark datasets.}
    \label{fig:dataset-images}
\end{center}
\end{figure}
\end{comment}

\begin{comment}
\begin{figure}[tbp]
\begin{center}
    \subfloat[CIFAR-100 \cite{krizhevsky2009learning}]{\includegraphics[ width=.31\columnwidth]{figures/CIFAR100_example_data2.png}\label{fig:images-CIFAR}} \hfill
    \subfloat[F-MNIST \cite{xiao2017fashion}]{\includegraphics[ width=.31\columnwidth]{figures/FashionMNIST_example_data2.png}\label{fig:images-FashionMNIST}} \hfill
    \subfloat[SVHN \cite{netzer2011reading}]{\includegraphics[ width=.31\columnwidth]{figures/SVHN_example_data2.png}\label{fig:images-SVHN}}
    \caption{Sample images from each of our benchmark datasets.}
    \label{fig:dataset-images}
\end{center}
\end{figure}
\end{comment}

\begin{table}[t!]
\caption{Characterization of CIFAR-100, F-MNIST, and SVHN.}
\begin{center}
\label{tab:datasets}
\begin{tabular}{@{}ll>{\centering\arraybackslash}m{1.6cm}>{\centering\arraybackslash}m{1.6cm}>{\centering\arraybackslash}m{1.9cm}@{}}
\toprule[2pt]
\multicolumn{2}{@{}l}{} & \textbf{CIFAR-100} & \textbf{F-MNIST} & \textbf{SVHN} \\ 
\midrule[1pt]
\multicolumn{2}{@{}l}{Number of samples} & 60,000 & 70,000 & 600,000 \\ \midrule[.5pt]
 & Training set   & 50,000 (83\%) & 60,000 (86\%) & 373,257 (62\%)\textsuperscript{\textdagger} \\ 
 & Validation set & 10,000 (17\%) & 10,000 (14\%) & 26,032 (4\%)\textsuperscript{\textdagger} \\ \midrule[1pt]
\multicolumn{2}{@{}l}{Number of classes} & 100 & 10 & 10 \\ \midrule[.5pt]
 & Samples per class & 600 & 7,000 & 6,500 to 19,000 \\ 
 & Balanced          & Yes & Yes & No \\ \midrule[1pt]
\multicolumn{2}{@{}l}{Number of channels}   & 3     & 1 & 3 \\ \midrule[1pt]
\multicolumn{2}{@{}l}{Sample size (pixels)} & 32$\times$32 & 28$\times$28 & 32$\times$32 \\ \midrule[1pt]
%\rule{0pt}{3cm}
\multicolumn{2}{@{}>{\flushleft\arraybackslash}m{2.5cm}}{Example image} &  \includegraphics[height=1cm]{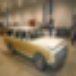}  & \includegraphics[height=1cm]{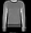} & \includegraphics[height=1cm]{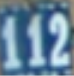}  \\ 
\bottomrule[2pt]
\end{tabular}
\end{center}
\textsuperscript{\textdagger}\footnotesize{SVHN also contains 531,131 samples that can be used as additional training data, which we did not use in our evaluation.}
\end{table}

\subsubsection{Neural Network Generation}
%We generate neural networks with randomized kernel, stride, and padding for both the convolutional and pooling layers. We randomize the width of the layers of the network, and we randomize the number of convolutional layers between 1 and 10. We add a pooling layer after each convolutional layer, and two fully connected layers at the end.
We train a set of diverse NNs on each of these datasets, where each NN is generated with randomly selected parameters.
We randomize the kernel, stride, and padding for convolutional layers (in ranges $[1, 5]$, $[1, 5]$, and $[0, 5]$, respectively) and pooling layers (in ranges $[1, 5]$, $[1, pool~kernel]$, and $[1, pool~kernel/2]$, respectively).%: kernel and stride for convolution are randomized between 1 and 5; padding for convolution is randomized between 0 and 5; kernel for pooling is randomized between 1 and 5; stride for pooling is randomized between 1 and the selected value for the pooling kernel; padding for pooling is randomized between 1 and half of the selected pooling kernel.
We also randomize the number of filters of each hidden convolutional layer (allowing a maximum of 400 for any given layer), number of convolutional layers (between 1 and 10), and the learning rate of the network (between \(10^{-6}\) and \(1+10^{-6}\)). We add a pooling layer after each convolutional layer, and two fully connected layers at the end.
Overall, we generate and train a set of 343 NNs on CIFAR-100, 505 NNs on F-MNIST, and 290 NNs on SVHN.

\subsection{Performance Estimation Engine Gain} \label{sec:savings}

%In answer to the first question we will discuss the percentage of training epochs saved using Curve Fit Predict as compared to MENNDL's early termination criterion, and we will demonstrate how this savings yields a theoretical increase in throughput
%To assess the savings of Curve Fit Predict, we begin by evaluating the percentage of training epochs saved by using Curve Fit Predict; then, we show the potential increase in throughput gained by these savings.
To answer first the question, \textit{``What savings are gained by our approach?"} we utilize %measure our gain in terms of 
two metrics---number of training epochs saved and throughput gain---for each of the three datasets. We compare the gain to state-of-the-art predictions using MENNDL~\cite{JohnstonICCS}.

\begin{figure*}[t!]
\begin{center}
    \subfloat[CIFAR-100]{\includegraphics[trim=9 10 11 10, clip, width=.33\linewidth]{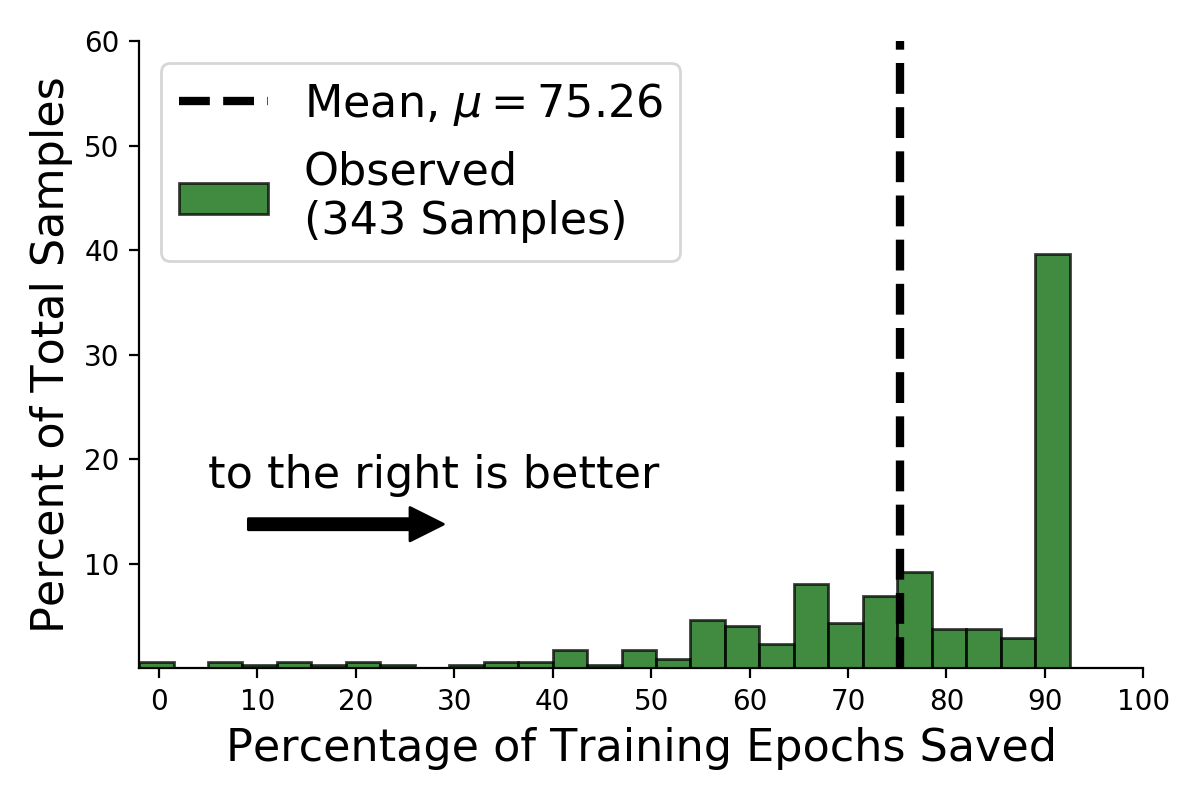}\label{fig:epochsSaved-CIFAR}}
    \subfloat[F-MNIST]{\includegraphics[trim=9 10 11 10, clip, width=.33\linewidth]{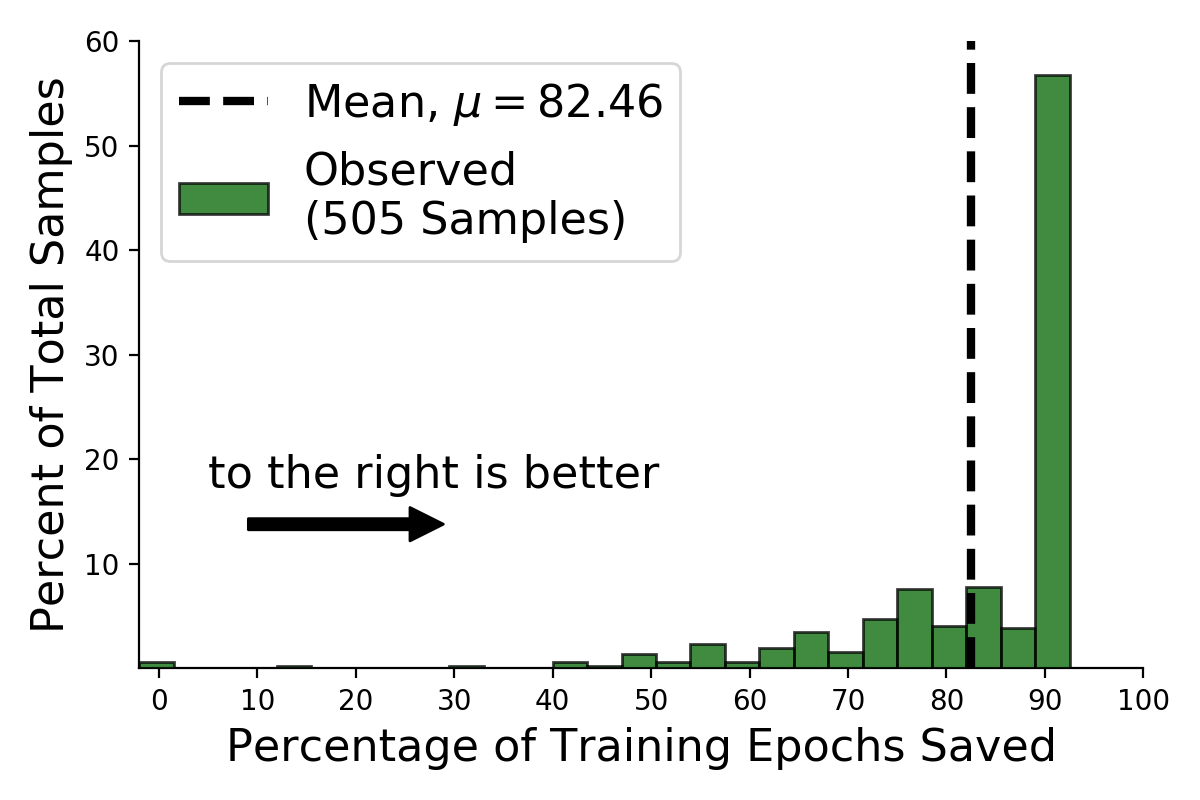}\label{fig:epochsSaved-FashionMNIST}}
    \subfloat[SVHN]{\includegraphics[trim=9 10 11 10, clip, width=.33\linewidth]{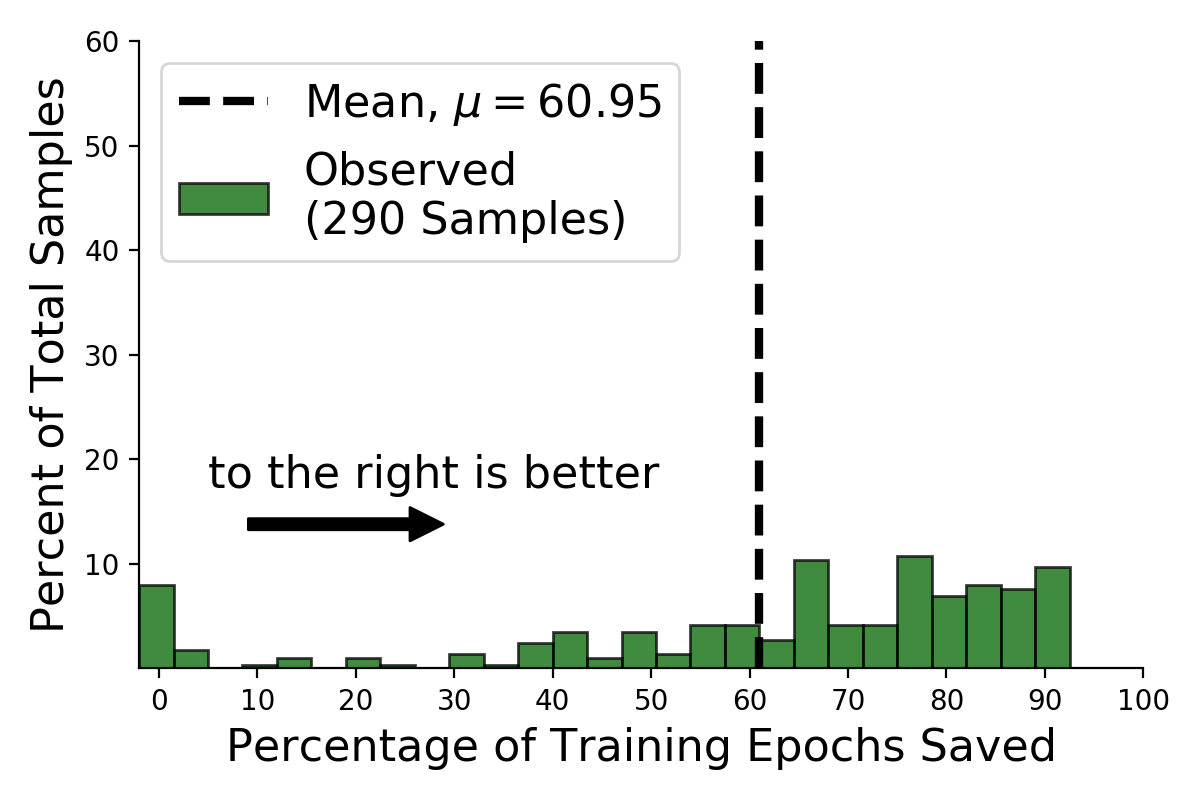}\label{fig:epochsSaved-SVHN}}
    \caption{Percentage of training epochs saved.}
    \label{fig:epochsSaved}
\end{center}
\end{figure*}

\subsubsection{Training Epochs Saved}

We compare the epoch at which \textit{PEng4NN} terminates the training of each NN model to the epoch at which MENNDL terminates training of that model using its built-in early termination criterion. MENNDL trains each NN for a maximum of 20 epochs, terminating training earlier than that if the network's minimum training loss has not decreased for at least 10 epochs. In contrast, \textit{PEng4NN} trains each NN for a maximum of 20 epochs, terminating training when the engine's prediction converges (as in Fig. \ref{fig:cfp_example}), avoiding additional training.

Figure \ref{fig:epochsSaved} depicts the distribution of NN models according to the percentage of training epochs our method saves for each NN, for our three datasets. Percentage of epochs saved is denoted on the \(x\)-axis and percent of total NN samples on the \(y\)-axis. The height of a rectangle denotes the percent of total samples that save the indicated portion of training epochs: taller rectangles to the right indicate that more samples save a larger percentage of epochs (i.e., there is a larger region of avoidable training for these samples). The dashed line indicates the mean percentage of training epochs saved.

For balanced datasets, such as CIFAR-100 and F-MNIST, \textit{PEng4NN}'s iterative process terminates as soon as the accuracy predictions converge to a value between 0\% and 100\%. We observe that 40\% to 50\% of models save more than 90\% of the training epochs on CIFAR-100 and F-MNIST (in Fig.~\ref{fig:epochsSaved-CIFAR} and Fig.~\ref{fig:epochsSaved-FashionMNIST}). For unbalanced datasets, such as SVHN, we describe in Section~\ref{sec:cfp} an additional condition that we require for prediction convergence for an NN with a \textit{never-learn prediction}. This delays convergence, and thus we see smaller savings on SVHN (in Fig.~\ref{fig:epochsSaved-SVHN}). Even so, we observe that nearly 40\% of models save 80\% or more of the training epochs on SVHN.

For all three datasets, we cut the mean number of training epochs needed to evaluate networks by well over half. We demonstrate respective mean savings of 75\%, 82\%, and 61\% of the training epochs on CIFAR-100, Fashion MNIST, and SVHN, as compared to MENNDL’s early termination criterion (indicated by the dashed lines in Figs.~\ref{fig:epochsSaved}a-c).

\subsubsection{Throughput Gain}

%\todo{Add here the theoretical comparison Travis is working on of MENNDL without CFP compared to MENNDL with CFP. Will that address the expected increase in throughput of the saved training epochs?}
%\textit{[BORROWED FROM THE PROPOSAL, FOR INSPIRATION: Figure~XX shows an example in which the surrogate model is extremely accurate and can stop training after 8 epochs (saving 60\% of the training time). Figure~YY shows how much time can be saved using the surrogate model. On average, we demonstrate a savings of approximately 65\% of the training time; approximately 84\% of the models can save 51\% or more of their computation time compared to fixed-length training of 20 epochs, showing the potential of the technique.]}

The ability of our performance estimation engine to predict final NN model accuracy early in the training process has immediate implications for NAS.
As shown in Figure~\ref{fig:epochsSaved}, we can anticipate reducing required training epochs by 60-80\% (mean savings depending on data set).
The reduction in training epochs for individual NNs increases the number of networks that can be explored and evaluated using the same amount of wall time and compute resources.% (reminiscent of weak scaling).

%Note that, for example on CIFAR-100 in Figure~\ref{fig:epochsSaved-CIFAR}, the mean savings is approximately 75\%, which corresponds to a throughput gain of approximately $4\times$, allowing $4\times$ the number of networks to be explored in the same amount of wall time with the same computational resources.

For example, on CIFAR-100 in Fig.~\ref{fig:epochsSaved-CIFAR}, the mean savings is approximately 75\%; this means \textit{PEng4NN} evaluates the models in \(\frac{1}{4}\) the computation needed by MENNDL. This corresponds to a throughput gain of about $4\times$, allowing $4\times$ the number of networks to be explored with the same computational resources. Similarly, \textit{PEng4NN} yields a throughput gain of $5\times$ on F-MNIST and $2.5\times$ on SVHN.
%Or if we need more detail about computing throughput for F-MNIST and SVHN, we can add back in the following sentences:
%For F-MNIST, the mean savings is approximately 80\%; this means we evaluate the models with \frac{1}{5} the computation, a theoretical throughput gain of $5\times$. For SVHN, the mean savings is approximately 60\%---using \frac{2}{5} the computational resources required by MENNDL, a theoretical throughput gain of $2.5\times$
Exploring more network structures % (larger problem size) 
gives the NAS additional opportunity to find better NN models. %improved structures (i.e., better NN models).
Alternatively, if the NAS had a fixed problem size (i.e., a set number of models to explore), then one could evaluate those models using fewer computational resources.
%in $\frac{1}{4}$ the time originally required (reminiscent of strong scaling) or 

\subsection{Performance Estimation Engine Accuracy} \label{sec:accuracy}
%To answer the second question, we study two metrics: overlap, which represents the proportion of the ground truth best models that are also in the set of predicted best models, and accuracy distribution, which shows how similar the neural network accuracies of the ground truth best models and predicted best models are.
We now address the second question, \textit{``How accurate are the predictions made by our approach?"} Recall that the NAS process involves generating models, evaluating their performance, selecting the best models at each step, and using these best models to inform the next generation of models (Section \ref{sec:intro}).
\textit{PEng4NN} reports accuracy estimates for all the generated NNs (see Section 
\ref{sec:cfp}), and the NAS uses these estimates to select the best models. 
This means it is important for \textit{PEng4NN} to accurately identify the best NN models from among the set generated by the NAS, as these are the models that are used to create the next generation of models. Thus, in evaluating our prediction accuracy, we compare the ground truth best x\% of all our generated models with the predicted best x\% of models as identified by \textit{PEng4NN}. We use \textit{PEng4NN}'s performance estimates for all the NNs to create a set of our engine's predicted best x\% of models. In our tests, in order to identify the ground truth best x\% of models, we allow the NNs to continue training and validating for a full 20 epochs, even after \textit{PEng4NN} has reported accuracy estimates for them. The models that achieve the highest validation accuracy are the ground truth best models.

We use two metrics to measure the accuracy of \textit{PEng4NN}'s predictions: \textit{overlap}, which represents the proportion of the ground truth best models that are also in the set of predicted best models, and \textit{accuracy distribution}, which depicts the similarity between the NN accuracy values of the ground truth and predicted best models.

\subsubsection{Overlap}

\begin{figure}[tbp]
\begin{center}
    \includegraphics[trim=0 15 0 0, clip,width=0.6\linewidth]{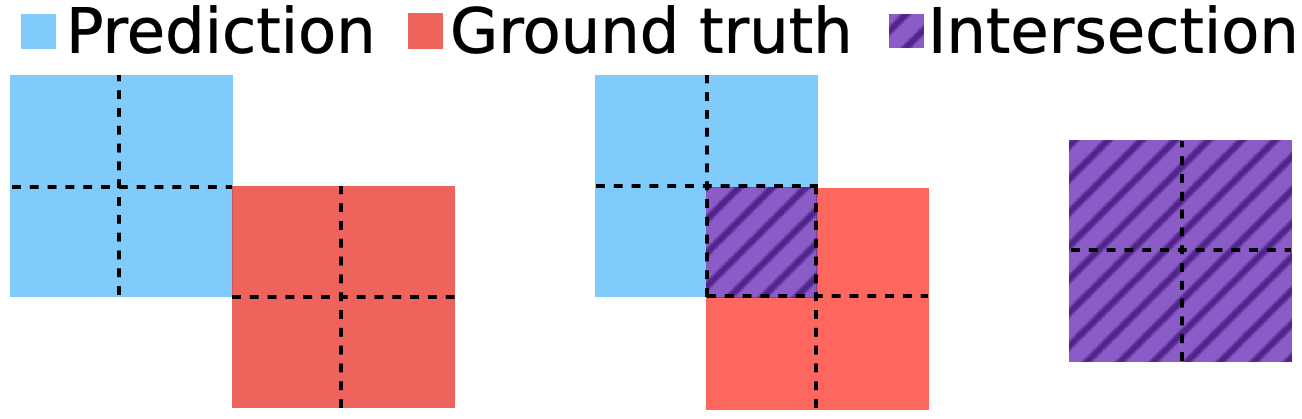}
    \caption{Overlap metric examples. \textit{Left}: overlap = 0, \textit{Center}: overlap = 1/4, \textit{Right}: overlap = 1}
    \label{fig:overlap}
\end{center}
\end{figure}

\begin{comment}
    \begin{figure}[tbp]
    \begin{center}
        \subfloat[\textit{Left}: overlap = 0, \textit{ Center}: overlap = 0.25, \textit{ Right}: overlap = 1]{\includegraphics[width=0.7\linewidth]{figures/overlapDefinition.png}\label{fig:overlap-a}} \\ %\qquad
        %\includegraphics[width=0.5\linewidth]{figures/overlapDefinition.png}
        %\caption{\textit{Left}: overlap = 0, \textit{Center}: overlap = 1/4, \textit{Right}: overlap = 1}
        \subfloat[\(|\text{ground truth}| = 10,\newline |\text{intersection}| = 8, \text{ overlap} = 0.8\)]{\includegraphics[width=0.7\linewidth]{figures/overlapExample.png}\label{fig:overlap-b}}
        \caption{Overlap metric examples.}
        \label{fig:overlap}
    \end{center}
    \end{figure}
\end{comment}

Overlap quantifies how many of the ground truth best models appear in the predicted best models and is defined as \(\frac{|\text{ground truth }\cap\text{ prediction}|}{|\text{ground truth}|}\). 
The overlap value ranges between 0 and 1; the closer to 1, the better. For example, \(\text{overlap} = 0.8\) would mean that \(8/10\) of the ground truth set is overlapping with the prediction set. One could trivially maximize the overlap by simply predicting more best models and allowing additional false positives. We avoid this by ensuring that the number of our predicted best models is always the same as the number of ground truth best models. See Figure \ref{fig:overlap} for simple illustrative examples of this metric with overlap values equal to 0, 25\%, and 100\%. 

Table \ref{tab:overlap-table} shows the overlap between \textit{PEng4NN}'s set of the predicted best \(x\)\% of NN models and the ground truth best \(x\)\% of NN models for different values of \(x\) and our three datasets. Across these datasets, for \(x\) values ranging from 10\% to 30\%, the overlap of ground truth best models and predicted best models lies between \(0.74\) and \(0.97\). 

\begin{figure*}[tbp]
\begin{center}
    \subfloat[CIFAR-100]{\includegraphics[trim=9 10 11 10, clip,width=.333\linewidth]{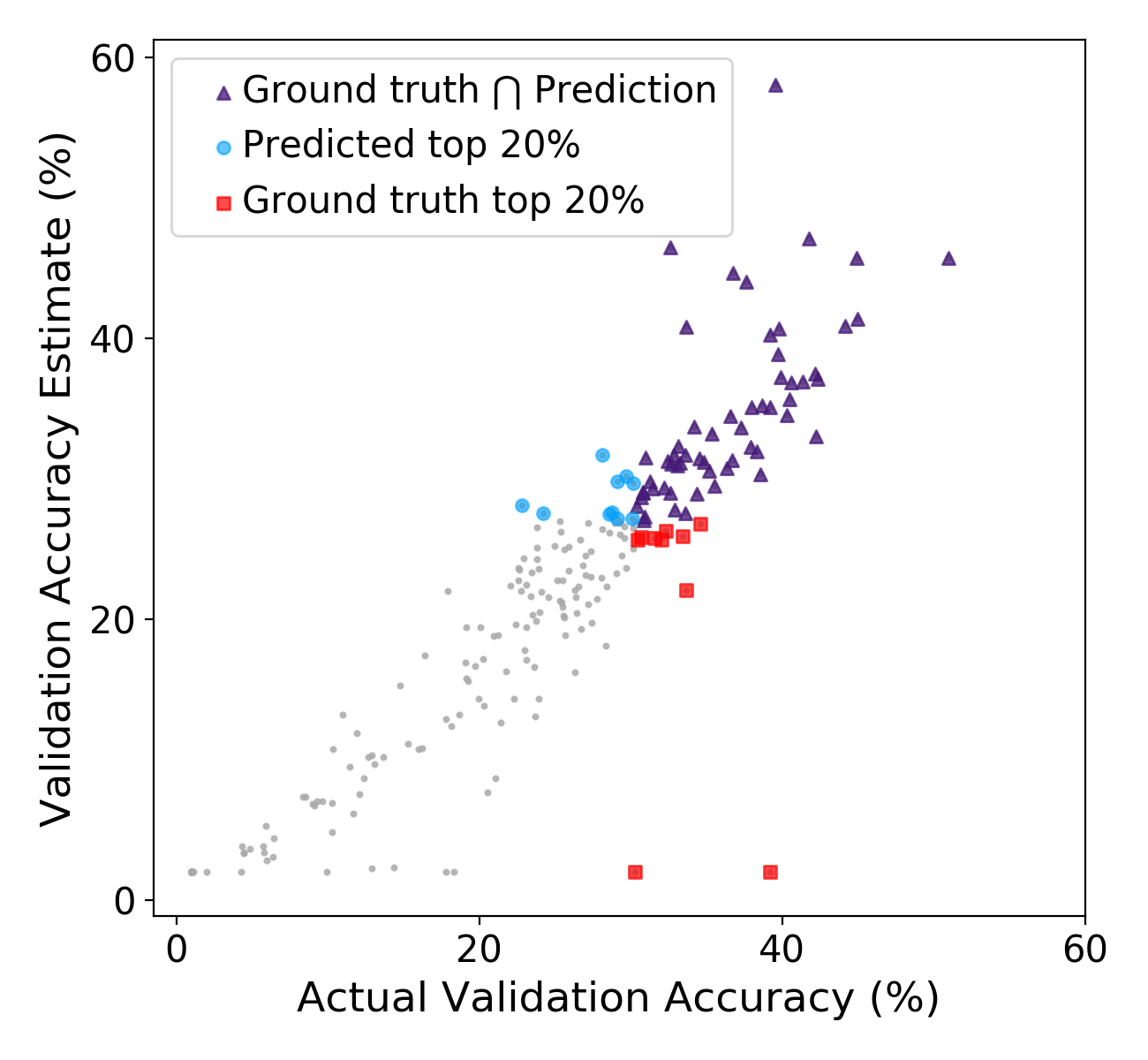}\label{fig:overlap20-CIFAR}}\hfill
    \subfloat[F-MNIST]{\includegraphics[trim=9 10 11 10, clip,width=.333\linewidth]{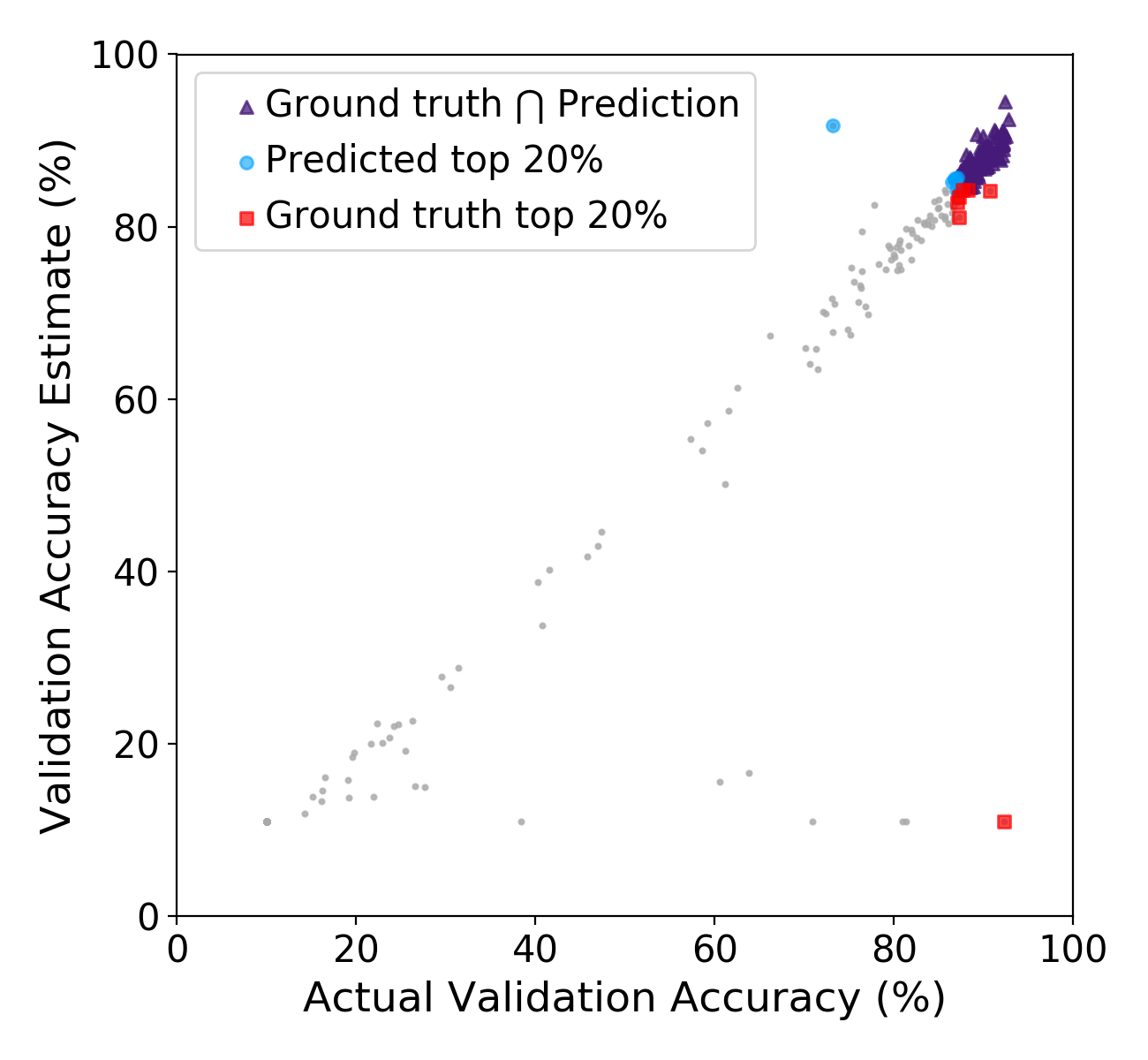}\label{fig:overlap20-FashionMNIST}}\hfill
    \subfloat[SVHN]{\includegraphics[trim=9 10 11 10, clip,width=.333\linewidth]{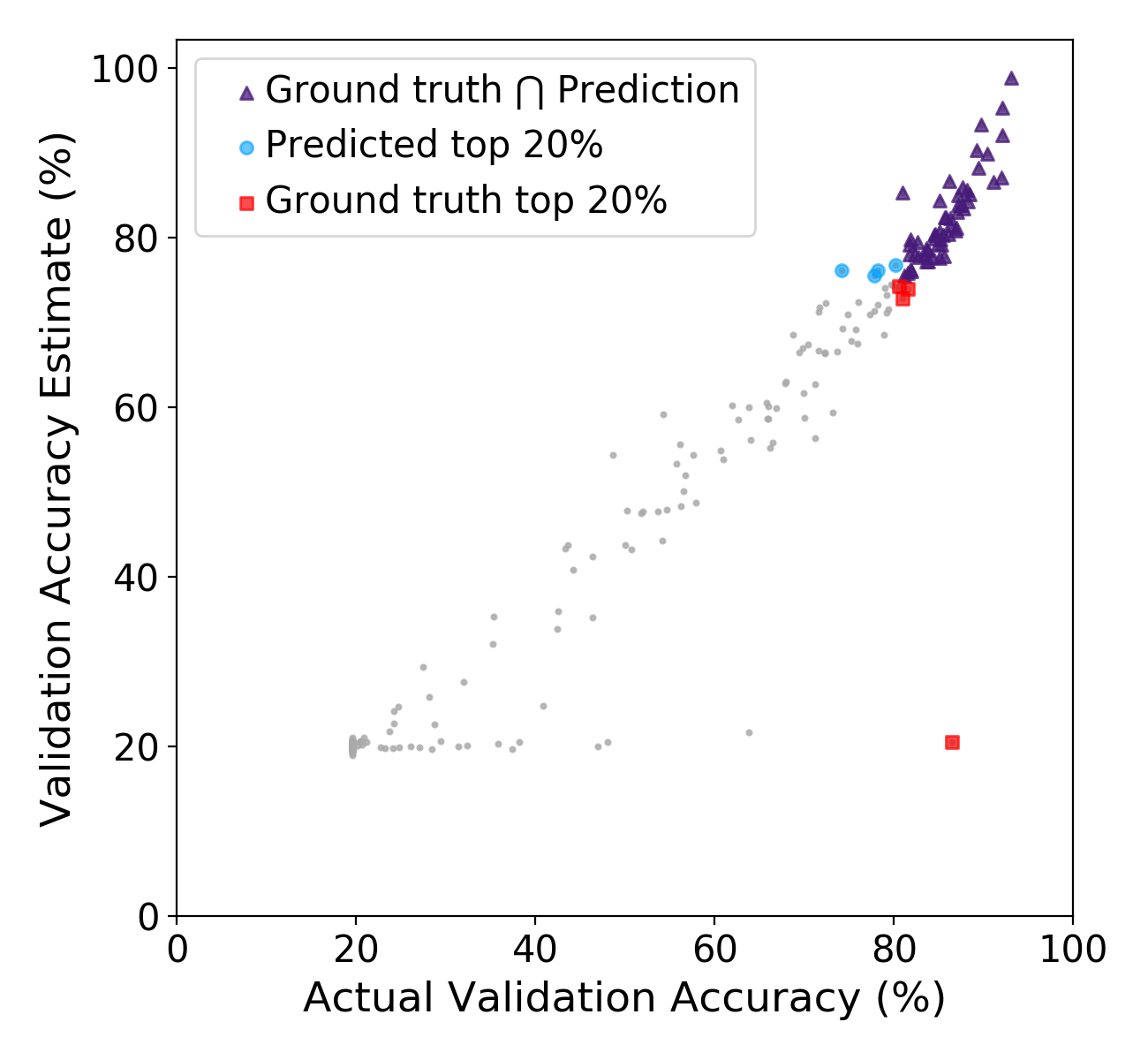}\label{fig:overlap20-SVHN}}
    \caption{Accuracy estimate and actual accuracy for all neural networks, indicating the intersection between ground truth best 20\% of models and predicted best 20\% of models.}
    \label{fig:overlap20}
\end{center}
\end{figure*}

\begin{figure}[tbp]
\begin{center}
    \subfloat[CIFAR-100]{\includegraphics[width=.33\linewidth]{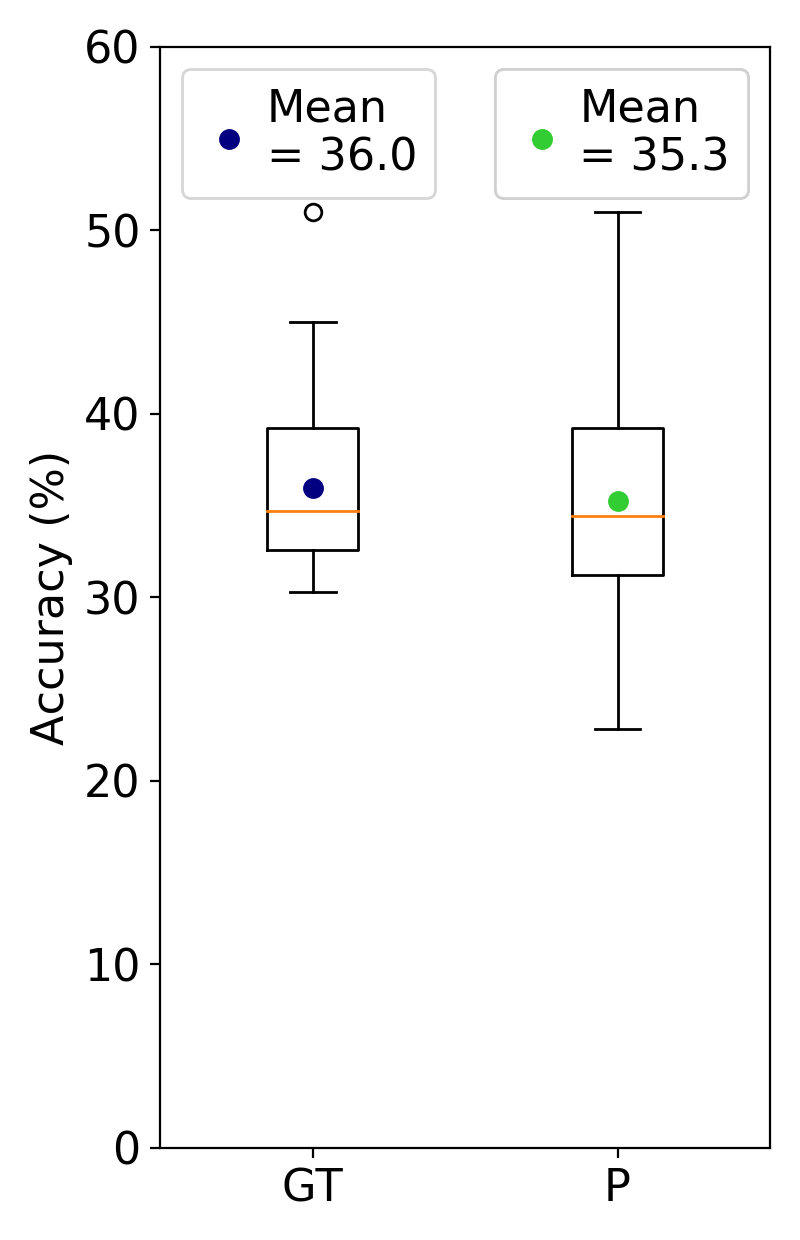}\label{fig:accRange-CIFAR}}
    \subfloat[F-MNIST]{\includegraphics[width=.33\linewidth]{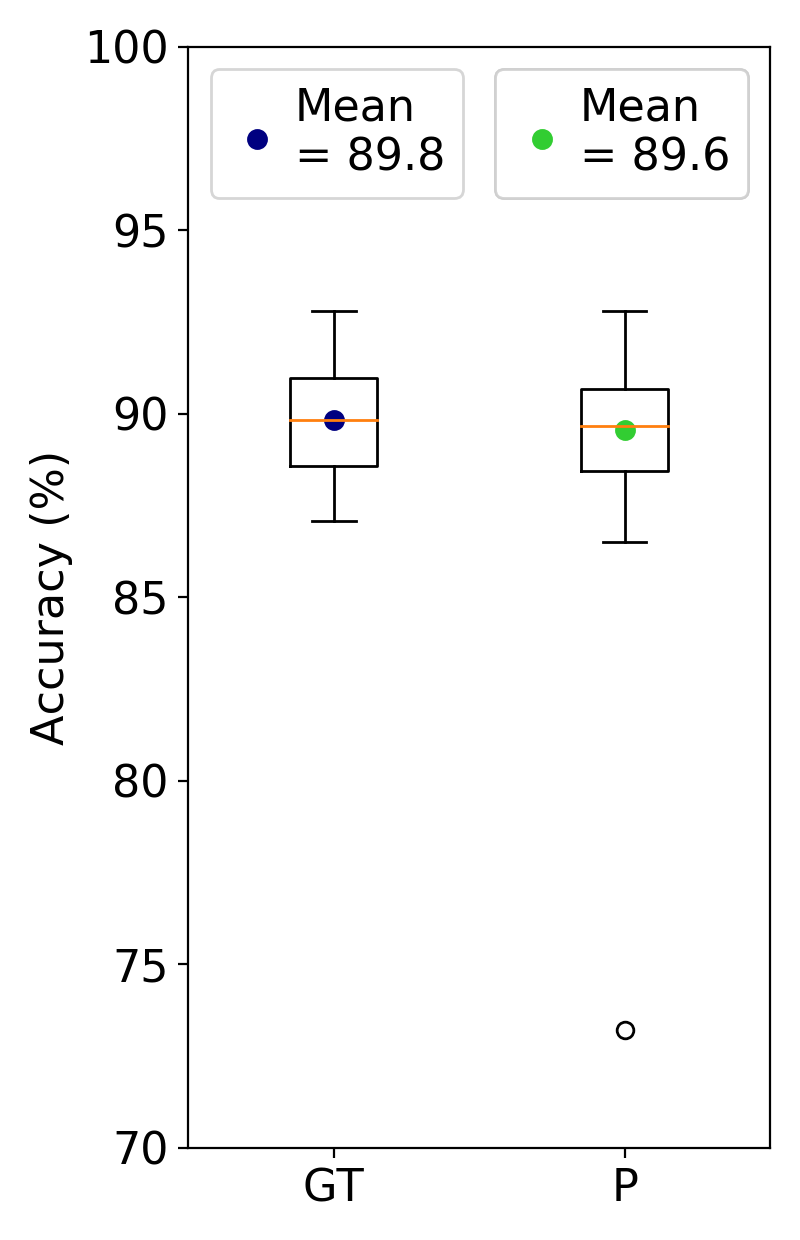}\label{fig:accRange-FashionMNIST}}
    \subfloat[SVHN]{\includegraphics[width=.33\linewidth]{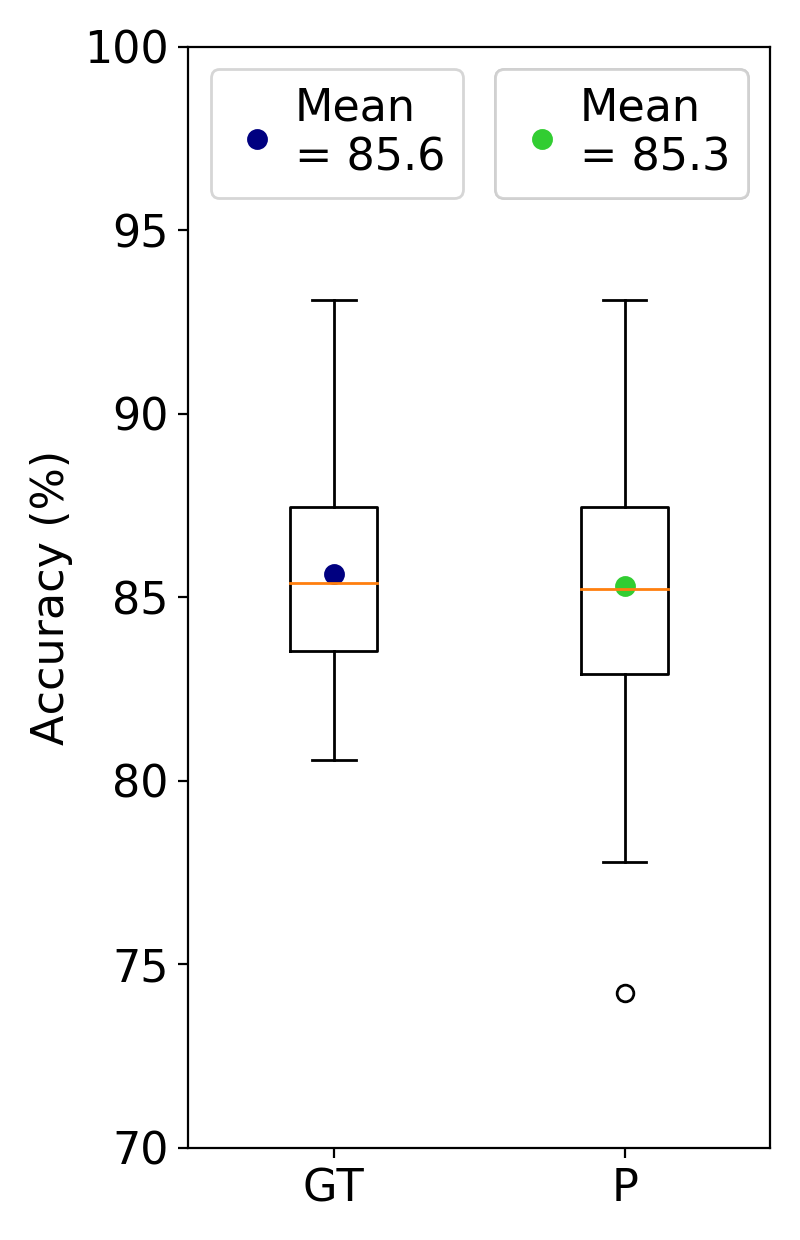}\label{fig:accRange-SVHN}}
    \caption{Network accuracy distribution of the top 20\% of models.\newline GT = ground truth top 20\%, P = predicted top 20\%}
    \label{fig:accRange}
\end{center}
\end{figure}

Figure \ref{fig:overlap20} corresponds to the second row of Table \ref{tab:overlap-table} and depicts, for each dataset, the maximum accuracy estimates and actual maximum accuracy values of all NN models, indicating the ground truth and predicted best 20\% and their intersection. Actual maximum accuracy is on the \(x\)-axis; the estimation engine's accuracy estimates are on the \(y\)-axis. The purple triangles depict all NNs that are in both the predicted best 20\% of models and the ground truth best 20\% of models. The blue circles depict NNs that are in the predicted best 20\% of models and not in the ground truth best 20\% of models. Last, the red squares depict NNs that are in the ground truth best 20\% of models but not in the predicted best 20\% of models. Because the \(x\)-axis indicates actual maximum accuracy, the most accurate models are the furthest to the right of the graph, and the least accurate models are the furthest to the left. We observe that the most accurate models from the set of ground truth best models lie in the intersection with the predicted best models, and the ground truth best models that lie outside of the intersection with the predicted best models tend to be the least accurate ones. 

Finally, in the lower left corner of the graph, we see NN models with very low actual accuracy and predicted accuracy; these are models that never learn (i.e. whose accuracy never increases). We see also that there are some outliers parallel to the \(x\)-axis that have varying levels of actual accuracy but are predicted to have very low accuracy. Validation accuracy for these models does not increase for several epochs. Part of our ongoing work involves studying ways to distinguish between NNs that never learn and NNs that do not learn for many epochs but eventually begin to learn (the outliers).

\subsubsection{Accuracy Distribution}

% We assess the similarity between NN accuracy values of the ground truth and predicted best models by their mean NN accuracy and NN accuracy distributions. %mean network accuracy and network accuracy distributions of the ground truth best models and predicted best models. 
We compare the ground truth and predicted best models with the mean and distribution of the NN accuracy values.
Table \ref{tab:accDiff-table} shows the difference between the mean accuracy values of the ground truth %best \(x\%\) of models and the mean accuracy of the 
and predicted best \(x\%\) of models for \(x=10,20,30\). In all cases, the mean network accuracy of the ground truth best models is very close to the mean network accuracy of the predicted best models: their difference ranges from 0.2 to 0.83 percentage points across all three image datasets and \(x\) values. 

Figure \ref{fig:accRange} corresponds to the second row of Table \ref{tab:accDiff-table} and depicts the NN accuracy distributions of the ground truth best 20\% of models and of the predicted best 20\% of models for each of our three datasets. %For each dataset, the actual best network is correctly included in the prediction set. Furthermore, 
The boxplot shows the 5th, 25th, 50th (red line), 75th and 95th percentiles. The dot represents the mean. Note that the CIFAR-100 figure has a different scale than the F-MNIST and SVHN figures. 
We observe that for each dataset, the NN accuracy distributions of the ground truth best models and of the predicted best models are comparable. In fact, for all three datasets, the mean NN accuracy for the ground truth best 20\% networks lies within 0.7 percentage points of the mean for the predicted best 20\% networks. On CIFAR-100, the mean accuracy of the ground truth best models is 36.0\%, while the mean accuracy of the predicted best models is 35.3\%. On F-MNIST, the mean accuracy of the ground truth best models is 89.8\%, while the mean accuracy of the predicted best models is 89.6\%. Last, on SVHN, the mean accuracy of the ground truth best models is 85.6\%, while the mean accuracy of the predicted best models is 85.3\%.
%percentage network accuracy

\begin{comment}
\begin{table}[tbp]
\caption{Overlap of the actual best \(x\)\% of models and the predicted best \(x\)\% of models for each dataset. \textit{Note: the sets of actual and predicted best models have the same size.}}
\begin{center}
\label{tab:overlap-table}
\begin{tabular}{p{.5cm}p{1.5cm}|p{1.5cm}|p{1.5cm}} 
    & CIFAR-100 & F-MNIST & SVHN \\ \cline{2-4}
    \multicolumn{1}{l|}{10\%} & \multicolumn{1}{l|}{0.79} & \multicolumn{1}{l|}{0.74}  & 0.90\\ \cline{1-4}
    \rowcolor[HTML]{C0C0C0}
    \multicolumn{1}{l|}{20\%} & \multicolumn{1}{l|}{0.74} & \multicolumn{1}{l|}{0.87} & 0.87\\ \cline{1-4}
    \multicolumn{1}{l|}{30\%} & \multicolumn{1}{l|}{0.88} & \multicolumn{1}{l|}{0.97} & 0.95
\end{tabular}
\end{center}
\end{table}
\end{comment}

\begin{comment}
\begin{table}[tbp]
\caption{Difference in the mean accuracy values of the ground truth best x\% of models and predicted best x\% of models.}
\begin{center}
\label{tab:accDiff-table}
\begin{tabular}{p{.5cm}p{1.5cm}|p{1.5cm}|p{1.5cm}} 
    & CIFAR-100 & F-MNIST & SVHN \\ \cline{2-4}
    \multicolumn{1}{l|}{10\%} & \multicolumn{1}{l|}{0.83} & \multicolumn{1}{l|}{0.66} & 0.22\\ \cline{1-4}
    \rowcolor[HTML]{C0C0C0}
    \multicolumn{1}{l|}{20\%} & \multicolumn{1}{l|}{0.70} & \multicolumn{1}{l|}{0.26} & 0.33\\ \cline{1-4}
    \multicolumn{1}{l|}{30\%} & \multicolumn{1}{l|}{0.58} & \multicolumn{1}{l|}{0.20} & 0.27
\end{tabular}
\end{center}
\end{table}
\end{comment}

\subsection{Discussion}\label{sec:discuss}

Our experiments show that across diverse datasets, \textit{PEng4NN} reduces needed training epochs and achieves a throughput gain compared to state-of-the-art NAS while maintaining accuracy in the predicted best models. CIFAR-100, F-MNIST, and SVHN are representative datasets that show diverse values for key dataset attributes (such as samples per class and sample distribution per class). Many of these attributes have implications for dataset classification difficulty and the performance of NNs. The demonstrated gain and accuracy of \textit{PEng4NN} across all three of these datasets exemplifies our engine's general applicability for datasets with disparate properties and difficulty.

We observe empirically that unbalanced datasets like SVHN can exhibit additional delay in learning and validation accuracy noise compared to balanced datasets like CIFAR-100 and F-MNIST. To address this, we utilize an additional condition for convergence on unbalanced datasets. If accuracy is not increasing, this condition checks if the NN's minimum validation loss has decreased within recent epochs. If validation loss has attained a new minimum, this can be an indication that the network is learning slowly, and the condition prevents early convergence in this case.

Furthermore, when studying overlap of the ground truth and predicted best models on all three datasets, we observe outlier models that are predicted to have very low accuracy but have various higher actual accuracy values. These are models that initially do not learn, but eventually learn after a several-epoch delay. Augmenting \textit{PEng4NN} to differentiate between these outliers and models that never learn is part of our current work.

\begin{table}[t!]
\caption{Overlap of the actual best \(x\)\% of models and the predicted best \(x\)\% of models for each dataset. \textit{Note: the sets of actual and predicted best models have the same size.}}
\begin{center}
\label{tab:overlap-table}
\begin{tabular}{@{}cccc}
\toprule[2pt]
    & \textbf{CIFAR-100} & \textbf{F-MNIST} & \textbf{SVHN} \\ \midrule[1pt]
    \multicolumn{1}{c}{10\%} & \multicolumn{1}{c}{0.79} & \multicolumn{1}{c}{0.74}  & 0.90\\ 
    \rowcolor[HTML]{E0E0E0}
    \multicolumn{1}{c}{20\%} & \multicolumn{1}{c}{0.74} & \multicolumn{1}{c}{0.87} & 0.87\\
    \multicolumn{1}{c}{30\%} & \multicolumn{1}{c}{0.88} & \multicolumn{1}{c}{0.97} & 0.95 \\
\bottomrule[2pt]
\end{tabular}
\end{center}
\end{table}

\begin{table}[t!]
\caption{Difference in the mean accuracy values of the ground truth best x\% of models and predicted best x\% of models.}
\begin{center}
\label{tab:accDiff-table}
\begin{tabular}{@{}cccc}
\toprule[2pt]
    & \textbf{CIFAR-100} & \textbf{F-MNIST} & \textbf{SVHN} \\ \midrule[1pt]
    \multicolumn{1}{c}{10\%} & \multicolumn{1}{c}{0.83} & \multicolumn{1}{c}{0.66} & 0.22\\ 
    \rowcolor[HTML]{E0E0E0}
    \multicolumn{1}{c}{20\%} & \multicolumn{1}{c}{0.70} & \multicolumn{1}{c}{0.26} & 0.33\\ 
    \multicolumn{1}{c}{30\%} & \multicolumn{1}{c}{0.58} & \multicolumn{1}{c}{0.20} & 0.27 \\
    \bottomrule[2pt]
\end{tabular}
\end{center}
\end{table}

\section{Related Works}
\label{sec:related}

%expand the impact of any existing neural architecture search (NAS) engine (e.g., MENNDL---Multinode Evolutionary Neural Networks for Deep Learning developed at ORNL and supported by ASCR) that could serve as a plugin generation module
%MENNDL~\cite{young2017evolving,young2019evolving} is an example of NAS systems that performs systematic search for optimum neural architecture, which performs better than off-the-shelf architectures for a variety of scientific tasks. It generates optimum NN models on DOE's Oak Ridge National Laboratory (ORNL) Summit supercomputer using an asynchronous, scalable evolutionary algorithm and applies the algorithm to a variety of scientific tasks~\cite{patton2019exascale,patton2018167}. Despite the success of NAS systems such as MENNDL, major challenges still exist in deploying NAS engines (see~\cite{elsken2019neural} for a detailed survey).

Recent works have proposed different performance estimation strategies. The two most common approaches for truncating training are to train for a fixed but significantly reduced number of epochs (e.g., 20 instead of 100s) \cite{real2017largescale,zoph2017neural} or to train until either the maximum accuracy or minimum loss has not improved for a user-specified amount of time (e.g., if the maximum observed accuracy does not improve for 5 epochs). Such methods are commonly included in NN software like Keras\footnote{See \texttt{https://keras.io/api/callbacks/early\_stopping/}}. % More references: ,zoph2017learning,real2018regularized
Limited research exists in the area of predicting NN performance \cite{domhan2015speeding, klein2017learning, baker2017accelerating, swersky2014freeze}. Some methods target hyperparameter search, which involves finding the best hyperparameter configuration for a given human-designed network but does not explore different architectures \cite{domhan2015speeding}. Domhan et al. \cite{domhan2015speeding} and Klein et al. \cite{klein2017learning} use probabilistic methods involving computationally expensive Markov Chain Monte Carlo sampling. The approach of Swersky et al. \cite{swersky2014freeze} automatically pauses and restarts training of models based on the predicted trajectory of the loss curves. Baker et al. \cite{baker2017accelerating} extract features from NNs to use in training a series of regression models to estimate performance. 

The solution presented in this paper amplifies these approaches by introducing two key components. The first one is the core accuracy predictive function that provides accurate estimations of the performance of an NN in terms of its maximum accuracy. The second is the prediction analyzer that ensures the stability of these estimations. We design these components to generate performance estimates agnostic to the NN model and the classification dataset in an effective way. 

\section{Conclusions}
\label{sec:conc}

This work introduces \textit{PEng4NN}, an engine that decouples the search and estimation strategies of NAS workflows. It generates accurate performance estimates for NN models and assists NAS strategies in the exploration of the neural architecture search space. Our solution is instrumental for NAS efficiency and throughput: it reduces the training epochs required by 60\% to 80\% on average, resulting in an increase in the throughput of explored networks of \(2.5\times\) to \(5\times\). Future work will leverage \textit{PEng4NN} in state-of-the-art NAS engines to demonstrate how it can increase the efficient interleaving of training, validation, and scientific predictions and analyses in analytics-driven workflows.

\section*{Acknowledgment}
This research used resources of the Oak Ridge Leadership Computing Facility at the Oak Ridge National Laboratory, which is supported by the Office of Science of the U.S. Department of Energy under Contract No. DE-AC05-00OR22725.

\bibliographystyle{IEEEtran}
\bibliography{references}

\end{document}